\documentclass[lettersize,journal]{IEEEtran}
\usepackage{amsmath,amsfonts}
\usepackage{algorithmic}
\usepackage{algorithm}
\usepackage{array}
\usepackage[caption=false,font=normalsize,labelfont=sf,textfont=sf]{subfig}
\usepackage{textcomp}
\usepackage{stfloats}
\usepackage{url}
\usepackage{verbatim}
\usepackage{graphicx}
\usepackage{float}
\usepackage{cite}
\usepackage{booktabs}
\usepackage{multirow}
\usepackage{hyperref}
\begin{document}
	
\title{CurriculumLoc: Enhancing Cross-Domain Geolocalization through Multi-Stage Refinement}

\author{{Boni Hu, Lin Chen, Runjian Chen, Shuhui Bu, Pengcheng Han, Haowei Li}
	% <-this % stops a space
	\thanks{Boni Hu, Lin Chen, Shuhui Bu, Pengcheng Han, Haowei Li are with the College of Aeronautics, Northwestern Polytechnical University, Xian 710000, China. (e-mail: huboni@mail.nwpu.edu.cn, npuchenlin@foxmail.com, bushuhui@nwpu.edu.cn, hanpc1125@mail.nwpu.edu.cn, 769292505@qq.com) 
	
	Runjian is with HKU-MMLab, University of Hong Kong. (e-mail: rjchen@connect.hku.hk)

	Corresponding author is Shuhui Bu
}}

\markboth{IEEE transactions on geoscience and remote sensing,Vol., No., November~2023}%
{Shell \MakeLowercase{\textit{et al.}}: A Sample Article Using IEEEtran.cls for IEEE Journals}


\maketitle

\begin{abstract}
Visual geolocalization is a cost-effective and scalable task that involves matching one or more query images, taken at some unknown location, to a set of geo-tagged reference images. 
Existing methods, devoted to semantic features representation, evolving towards robustness to a wide variety between query and reference, including illumination and viewpoint changes, as well as scale and seasonal variations.
However, practical visual geolocalization approaches need to be robust in appearance changing and extreme viewpoint variation conditions, while providing accurate global location estimates.
Therefore, inspired by curriculum design, human learn general knowledge first and then delve into professional expertise. We first recognize semantic scene and then measure geometric structure.
Our approach, termed \emph{CurriculumLoc}, involves a delicate design of multi-stage refinement pipeline and a novel keypoint detection and description with global semantic awareness and local geometric verification. 
We rerank candidates and solve a particular cross-domain perspective-n-point (PnP) problem based on these keypoints and corresponding descriptors, position refinement occurs incrementally. 
% resulting in aforementioned desirable characteristics of a practical visual geolocalization solution.
% Specifically, a novel vision transformer, \emph{Swin-Descriptors}, is designed to detect and describe keypoints, incorporating a symmetric Encoder-Decoder Swin Transformer with shifted window attention mechanism and soft detection score strategy.
%Reranking candidates and solving a particular cross-domain perspective-n-point (PnP) problem based on these keypoints and corresponding descriptors, position refinement occurs incrementally.
%
The extensive experimental results on our collected dataset, \emph{TerraTrack} and a benchmark dataset, \emph{ALTO}, demonstrate that our approach results in the aforementioned desirable characteristics of a practical visual geolocalization solution. Additionally, we achieve new high recall@1 scores of 62.6\% and 94.5\% on ALTO, with two different distances metrics, respectively.
Dataset, code and trained models are publicly available on \href{https://github.com/npupilab/CurriculumLoc}{https://github.com/npupilab/CurriculumLoc}.

% This paper introduce a new UAVs localization framework based on locally-global geometric verification after retrieval. Retrieval based on NetVLAD, We further propose a deep local feature point extract network, then using geometric verification to rerank global feature retrieval results and using perspective-n-point to compute the precise location. While we This locally-global features are highly invariant to both appearance (year, season, and illumination) and viewpoint (translation and rotation) changes in the always changing world.
%

\end{abstract}


\section{Introduction}
%
% When humans determine their location, they not only consider global semantic context information to recognize scene， but also pay attention to the multi-view geometric relationships contained in useful images for a more accurate result. 可以用在intro
%
\begin{figure}[!t]
	\centering
	\includegraphics[width=3.5in]{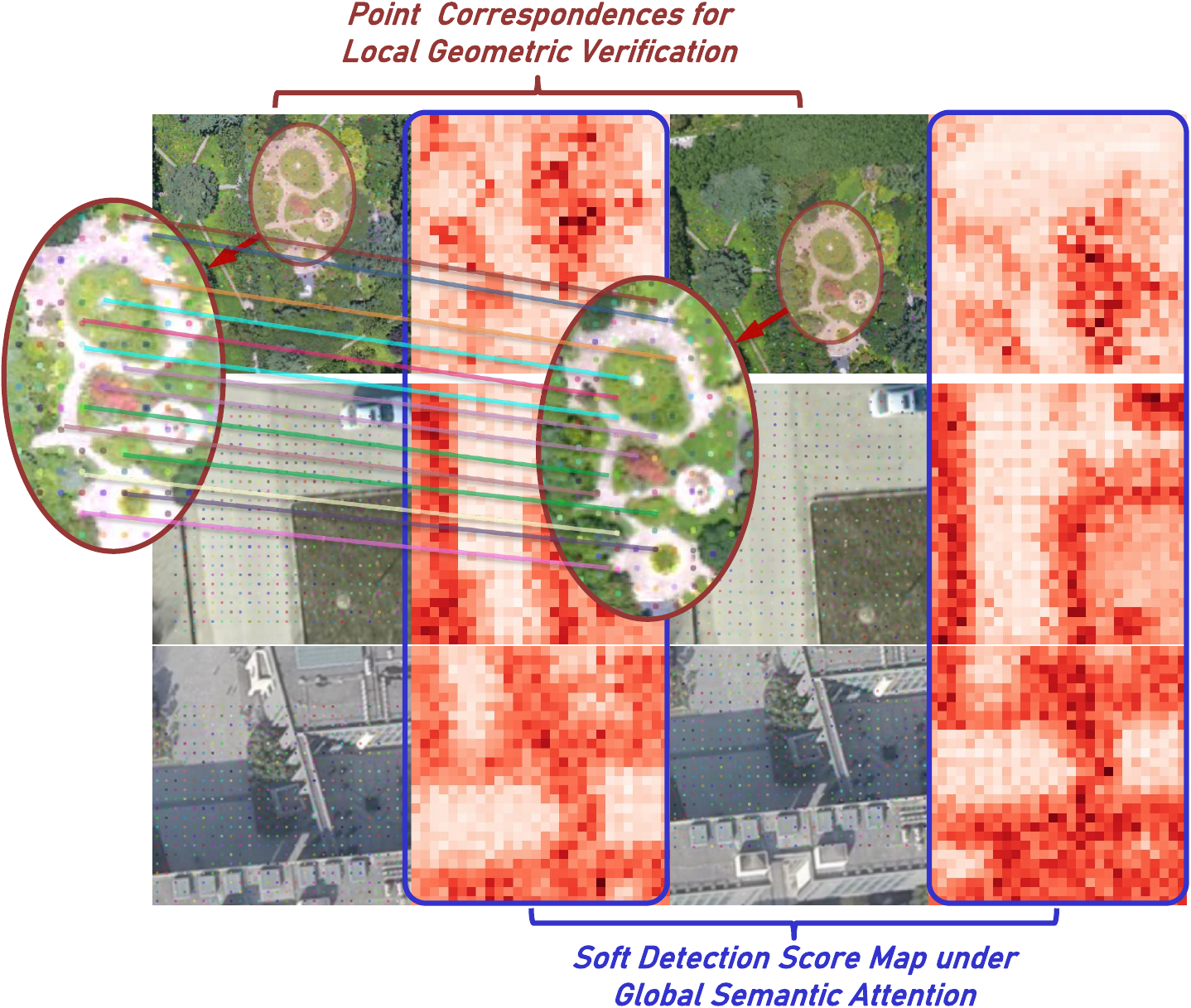}
	\caption{\textbf{Visualization of pixel correspondences supervision and soft detection score of matched images during training.} White represents low soft-detection scores while red signifies higher ones. The training lowers the soft-detection scores on repetitive structures (e.g. ground, floor, walls) while it enhances the score on more distinctive points. And during training soft detection scores are optimized under the supervision of pixel correspondences.}
	\label{fig1}
\end{figure}
\IEEEPARstart{A}{} high quality, view aware \footnote{Images captured with an intention to localize with a wide view of the surrounding.} image often captures sufficient information to uniquely represent a location \cite{soft_2020} . Meanwhile, it is convenient to capture aerial images, with low cost and high adaptability camera. Therefore, it is not surprising that we use vision as the primary source of information for UAVs localization, navigation, and exploration in remote sensing. 

Currently, vision-based localization is tackled either with structure-based methods, such as Structure-from-Motion (SfM) \cite{SfM_2017_torsten, rajamohan2022image,  Sattler_Leibe_Kobbelt_2017, nota2022improving} and Simultaneous Localization and Mapping (SLAM) \cite{Bresson_Alsayed_Yu_Glaser_2017, taheri2021slam, chen2022overview, gupta2022simultaneous}, or with retrieval-based approaches \cite{Arandjelovic_Gronat_Torii_Pajdla_Sivic_2018, PatchNetVLAD, zhang2023gcg, xu2023transvlad, delg_2020}. 
Structure-based methods estimate relative poses based on precise 3D-2D point correspondences, which can only output relative positions rather than geolocations, and the estimation error will increase with the accumulation of distance \cite{long_term_2022}. 
The application of geolocalization in remote sensing is limited primarily because of the high computational demand \cite{Xu_Pan_Du_Li_Jing_Wu_2018}. This limitation is usually alleviated using additional sensory systems such as Global Positioning System (GPS) and landmarks \cite{Xin_Jiang_Zou_2019}. However, the loss of GPS signal is inevitable due to various unpredictable factors in real-world flight missions. 
Furthermore, most existing structure-based methods focus on the accurate local descriptions of individual keypoints, which neglect their relationships established from global awareness \cite{superpoint_Rabinovich_2018, d2Net_2019}. It's easy to generate mismatches due to appearance changes such as lighting, viewpoint, and weather, and even cause positioning failures, especially in cross-domain tasks. There is a well-known trade-off between discriminative power and invariance for local descriptors \cite{long_term_2022}.

Retrieval-based geolocalization is an appealing alternative that estimates the location where a given photo was taken by comparing it with a large database of geo-tagged reference images. 
Recent developments in learning image features for object detection and semantic segmentation have made image retrieval a viable method for localization \cite{liu2022swin, he2022masked, kirillov2023segment, domainfeat_2023}. Its main goal is to design high-level semantic feature representation algorithms that are invariant to both appearance (year, season, and illumination) and viewpoint (translation and rotation) changes in the always changing world, like human instinct \cite{SouravGarg2021WhereIY}. However, due to the lack of geometric transformation, retrieval-based geolocalization cannot provide accurate pose and associated location. 
Moreover, it needs to deal with additional challenges in remote sensing, primarily stemming from the domain disparities between the query UAV images and the reference satellite images.
%And facing additional challenges in remote sensing due to the differences between query UAV and reference satellite images from different domains.
% When human perceive an image, they initially quickly recognize familiar visual scenes, subsequently turn their attention to some key regions for spatial analysis, and finally arrive at a more precise location. 

In this work, we rethink the visual geolocalization problem from the lens of robust human perception and precise geometry measurement. 
We first adopt a general retrieval approach to obtain the top-$k$ candidates of the reference database, which are the nearest neighbors to the current aerial image in the semantic space of global representations. These global descriptors represents the invariant features related to locations, initial retrieval in this feature space helps to ignore the interference of appearance changes such as lighting, season, weather, etc. on localization robustness. 

Besides, we introduce a novel symmetric Swin Transformer with skip connections and shift window attention for keypoints descriptors construction called \emph{Swin-Descriptors}. The motivation is to use the attention in Swin Transformer to enhance nonlocal awareness, thereby enhancing the invariance to appearance changes at the same location. 
Furthermore, to enhance the acquisition of salient and consistent keypoints and their corresponding descriptors representations, we devise a soft detection score to detect keypoints, which is derived from transformer extracted dense feature and is optimized by an extend ranking loss with point geometric verification. 
Our experiments show that descriptors construct from our \emph{Swin-Descriptors} are promising  in addressing the trade-off between discriminative power and invariance of local descriptors.

Besides, we rerank the initial candidates through the stereo matching relationships based on these keypoint descriptors, which significantly enhances the retrieval performance, particularly in terms of recall@1 and recall@5.
Moreover, by utilizing the discriminative power and invariance of these keypoints descriptors, we formulate a cross-domain PnP problem between the query UAV image and the corresponding candidate geotagged satellite images. We show that solving this PnP problem yields a more accurate geolocation estimate than the geotag obtained from reranked recall@1.
% We show that, through multi-stage location refinement, cross-domain visual geolocalization can remain robust in apperance changes between query UAV domain and reference satellite domain, and obtain accurate global positions instead of the geo-tag of nearest neighbor reference image.  
%In this paper, we begin by using retrieval and rerank approach to obtain the top-$k$ candidates, which are the nearest neighbors to the current aerial image in the semantic feature space. Subsequently, we establish 3D-2D correspondences for geometric verification.
%

In summary, our main contributions are:
\begin{itemize}
	\item 
	We propose a new pipeline of visual geolocalization called \emph{CurriculumLoc}, which measuring the cross-domain geometric co-visibility structure after obtaining a coarse location through an incremental retrieval, resulting in robustness and desirable accuracy of a practical visual geolocalization solution.
	
	\item By leveraging a novel symmetric hierarchical transformer with attention mechanism during dense feature extraction, and a soft detection score for keypoint detection derive from dense feature under pixel correspondences supervision, the keypoint undergo comprehensive exposure to both global contextual awareness and local geometric verification during the training process, as illustrated in Fig. \ref{fig1}.

	\item To evaluate the effectiveness of \emph{CurriculumLoc}, we create a publicly available dataset containing sequences of UAV and satellite imagery captured under challenging conditions for cross-domain geolocalization, named \emph{TerraTrack}.

\end{itemize}

\section{Related Work}
In this section, we discuss the recent researches that utilize structure-based methods or retrieval-based methods for visual localization and transformer architecture for vision tasks.
\subsection{Structure-based visual localization}
Existing structure-based localization works can broadly divided into: (i) Structure-from-Motion (SfM); (ii) Simultaneous Localization and Mapping (SLAM). 
Both of these, when not augmented with additional sensors, are capable of obtaining only relative position information. Specifically, SfM is primarily employed for offline detailed reconstruction, while SLAM focuses on real-time localization and mapping. Both rely heavily on precise relative pose estimation, which are greatly dependent on the quality of keypoint detection and description.

In early localization systems \cite{orbslam2_2017, gslam_2019}, conventional techniques like SIFT \cite{Lowe_1999_sift}, SURF \cite{bay2008speeded_surf} and ORB \cite{Konolige_Bradski_2011_orb} were extensively employed to characterize a small patch centered on a detected keypoint. However, it's evident that these manually designed features with out global semantic attention struggle with significant appearance variations. 
More recently, CNN-based methods have demonstrated superior performance. Some of these approaches have suggested a two-stage process where keypoints are initially detected based on local structures and subsequently described using a separate CNN \cite{lfNet_2018, Zhang_Rusinkiewicz_2018}. Notably, LIFT \cite{lift_2016} and LFNet \cite{lfNet_2018} present distinct two-stage algorithms for keypoint detection and patch description based on the detected keypoints. 
In contrast to the aforementioned two-stage methods, certain approaches advocate for a unified one-stage approach for keypoint detection and description. SuperPoint \cite{superpoint_Rabinovich_2018}, for instance, employs a self-supervised framework with a bootstrap training strategy to concurrently train a model for keypoint detection and descriptor extraction. On the other hand, R2D2 \cite{2019_r2d2} incorporates effective loss functions to assess both the repeatability and reliability of keypoints detection.
Furthermore, DomainFeat \cite{domainfeat_2023} enhances the extraction of robust descriptors and the detection of accurate keypoints by leveraging domain adaptation to learn local features.

Different from these prior works, we recommend the introduction of non-local awareness to include both hierarchical feature extraction with shift window attention mechanism and geometric verification, aiming to make the keypoints descriptors remarkable (easy to extract), invariant (not varying with rotation, translation, scaling and illumination changes) and accurate (accurate to measure).
\subsection{Retrieval-based visual localization}

\subsubsection{Global Descriptors}

Traditional global descriptors are usually obtained by aggregating local descriptors, such as Bag of Words (BoW), Fisher Kernal, and Vector of Locally Aggregated Descriptors (VLAD), have been used to assign visual words to images. 
Since the remarkable results of the NetVLAD \cite{Arandjelovic_Gronat_Torii_Pajdla_Sivic_2018} algorithm based on contrastive learning and adopting CNNs, researchers have been focusing on how to extract accurate and robust global descriptors to achieve retrieval localization.

Recent advances in global representation for retrieval cover a wide range of techniques and strategies. These include ranking loss-based learning \cite{Revaud_Almazan_Rezende_Souza_2019}, soft contrastive learning \cite{soft_2020}, innovative pooling methods \cite{Radenovic_Tolias_Chum_2019}, contextual feature reweighting \cite{Kim_Dunn_Frahm_2017}, large-scale retraining \cite{Warburg_Hauberg_Lopez-Antequera_Gargallo_Kuang_Civera_2020}, semantic-guided feature aggregation \cite{semantic2021}, 3D information integration\cite{structure_cue}, incorporation of additional sensors like event camera \cite{event2020}, sequence information \cite{delta_2020}, graph representation \cite{GCG-Net} and training with classification proxy \cite{GCG-Net,cosplace}.

Distinct from these retrieval methods only focuses on global representation of images, we argue that global descriptors are suitable for robust scene recognition to determine a rough location range, while exact location requires geometric motion model from accurate local correspondences.

\subsubsection{Local Region/Patch Descriptors}

In addition to global retrieval methods, some researchers have also been dedicated to learning task-specific patch-level features for place recognition. 
Patch-NetVLAD \cite{PatchNetVLAD} demonstrates that this two-stage retrieval strategy, global retrieval and then rerank based on local features correlation verification, can improve the accuracy of place recognition. 
Patch-NetVLAD \cite{PatchNetVLAD} exploit patch-level features from pretrained NetVLAD \cite{Arandjelovic_Gronat_Torii_Pajdla_Sivic_2018} residuals. Distinct from Patch-NetVLAD only considering aggregating local features, TransVLAD \cite{xu2023transvlad} design a attention-based feature extraction network with a sparse transformer, which both improves global contextual reasoning and aggregates a discriminative and compact global descriptor. %

Furthermore, we design an Encoder-Decoder hierarchical transformer, \emph{Swin-Descriptors}, to detect and describe keypoints. We fully leverage the geometric relationship between matched points to obtain more accurate rerank top-1, and return the final precise position by solving the camera motion model of UAV, rather than return the closest satellite image geo-tag.

\begin{figure*}[!t]
	\centering
	\includegraphics[width=7.0in]{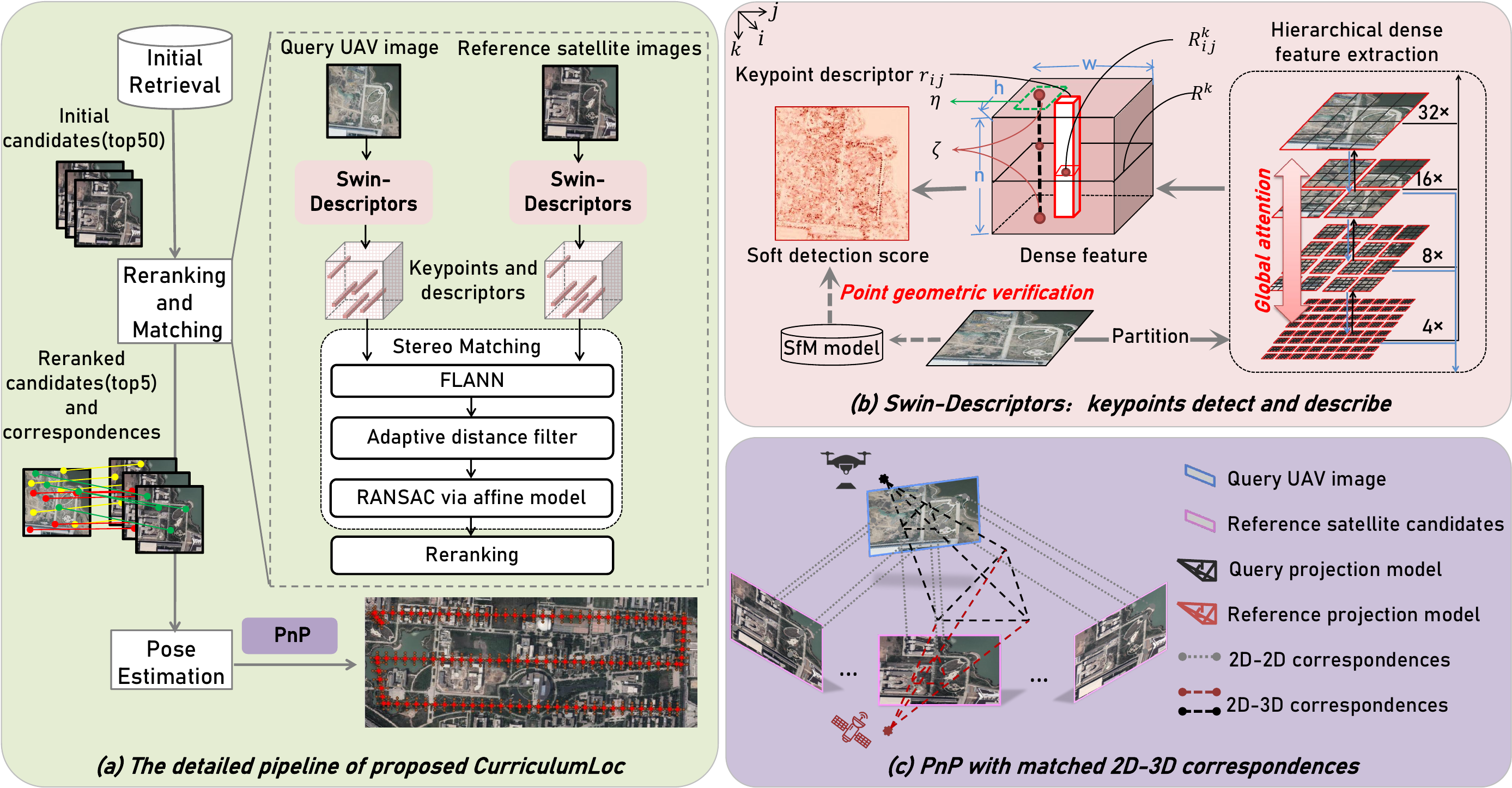}
	\caption{(a): The detailed pipeline of proposed \emph{CurriculumLoc}. (b): Details of \emph{Swin-Descriptors} in (a). (c): The schematic of cross-domian PnP in (a).}
	\label{outline}
\end{figure*}
\subsection{Attention for visual localization}

Attention mechanism, inspired by the human visual and cognitive systems, has been successfully employed applied to the learning of image-level global descriptors \cite{Tolias_Jenicek_Chum_2020, attention_2017,xu2023transvlad} for image retrieval tasks.
In these methods, the attention is used as the weight of local descriptors, and global descriptors are derived from local descriptors through weighted summation. 
Directly applying these descriptors lead to poor results due to the lack of supervision of local pixel correspondence \cite{2019_r2d2}. Attention mechanisms in these methods are optimized with the supervision of image-level, and it is not suitable for the pixel correspondence supervision. Moreover, existing attention mechanisms designed for location-based image retrieval is encouraged to focus on the most representative regions of the image rather than identifying suitable regions to maintain the location consistency and detecting invariant points for geometric matching.  

In contrast, we propose a novel soft detection score for attention computing and feature extraction to improve the keypoint geometric consistency and matching accuracy. On the one hand, our soft detection score can find remarkable points for matching (easily detect to inliers while distinguishable to outliers) through a SfM model supervision. On the other hand, our soft detection enforces the same response for the matched points between different domain query and reference images, so it can provide extra priors to improve the robustness in matching process under changes in the appearance of the same location. Meanwhile, we leverage the shift window attention in Swin Transformer for our symmetric transformer to maintain scale invariance. 

\section{Methodology}

Our approach begins by utilizing global semantic feature to obtain the top-$k$ candidates (where $k$=50) that are most likely to represent the same location as the given query image. Following this, we employ a novel keypoint descriptor construction method called \emph{Swin-Descriptors} to detect and describe keypoints in both query image and the top-$50$ candidates. Simultaneously, we rerank these reference images, and return the selection of the top-$5$ candidates along with their corresponding matched keypoint pairs.
In the end, by computing the PnP using matched keypoint pairs between the top-$5$ candidates and the current UAV image, we can recover the camera motion model of UAV, ultimately achieving a more accurate localization than relying solely on the geotag of the nearest satellite in the reference database.

This multi-stage approach enhances geolocalization accuracy and robustness while minimizing the additional computational cost associated with keypoint matching and PnP solving. An overview of the complete pipeline can be found in Fig. \ref{outline}(a).

\subsection{Initial Retrieval}
 
We use a lightweight VGG-16 \cite{Simonyan_Zisserman_2015} network cropped at the last convolutional layer and extend it with a NetVLAD \cite{Arandjelovic_Gronat_Torii_Pajdla_Sivic_2018} layer as implemented by \cite{PatchNetVLAD}, initializing the network with off-the-shelf ImageNet \cite{Fei_Fei_2009} classification weights, and retaining the first few layers of weights for general feature extraction.
Specifically, given an query image $ \mathcal{q}$, outputs a $H \times W \times D$ dimensional features into a $K \times D$ dimensional matrix by summing the residuals between each feature $x_i \in \mathbb{R}^D$ and $K$ learned cluster centers weighted by soft-assignment. Formally, for $N \times D$ dimensional features, let the global aggregation layer $f_{global}: \mathbb{R}^{N \times D} \rightarrow \mathbb{R}^{K \times D}$ be given by
\begin{equation}
\label{netvlad}
f_{global}(F)(j,k)=\sum_{i=1}^N \bar{a}_k(X_i)(x_i(j)-c_k(j)),
\end{equation}
where $x_i(j)$ is the $j^{th}$ element of the $i_{th}$ descriptor, $\bar{a}_k$ is the soft-assignment function and $c_k$ denotes the $k^{th}$ cluster center. After aggregation, the resultant matrix is then projected down into a dimensionality reduced vector using a projection layer $f_{proj}: \mathbb{R}^{K \times D} \rightarrow \mathbb{R}^{D_{proj}}$ by first applying intra (column) wise normalization, unrolling into a single vector, L2-normalizing in its entirety and finally applying PCA (learned on a training set) with whitening and L2-normalization. Refer to \cite{Arandjelovic_Gronat_Torii_Pajdla_Sivic_2018} for more details.

We use this global feature for initial retrieval and return top-$50$ reference satellite images. This step helps us quickly retrieve satellite images from a database that are geographically close to the query UAV image's location, primarily based on feature distance, even though it may contain erroneous matches. Subsequently, reordering is performed based on geometric verification through local point matching, which enhances accuracy while saving time.

\subsection{Swin-Descriptors}

Contrary to existing two-stage methods, which rerank global retrieval results by local semantic feature verification, such as PatchNetVLAD \cite{PatchNetVLAD}, we propose to preform dense semantic feature extraction to obtain a local representation that is enable to capture remarkable keypoints with global awareness and pixel level geometric supervision. As illustrated in Fig. \ref{outline}(b), our keypoints detection and description share the underlying representation, obtained by a hierarchical feature extractor with global contextual attention, called \emph{Swin-Descriptors}. 

\subsubsection{Dense Feature Extraction}

\begin{figure*}[!t]
	\centering
	\includegraphics[width=6in]{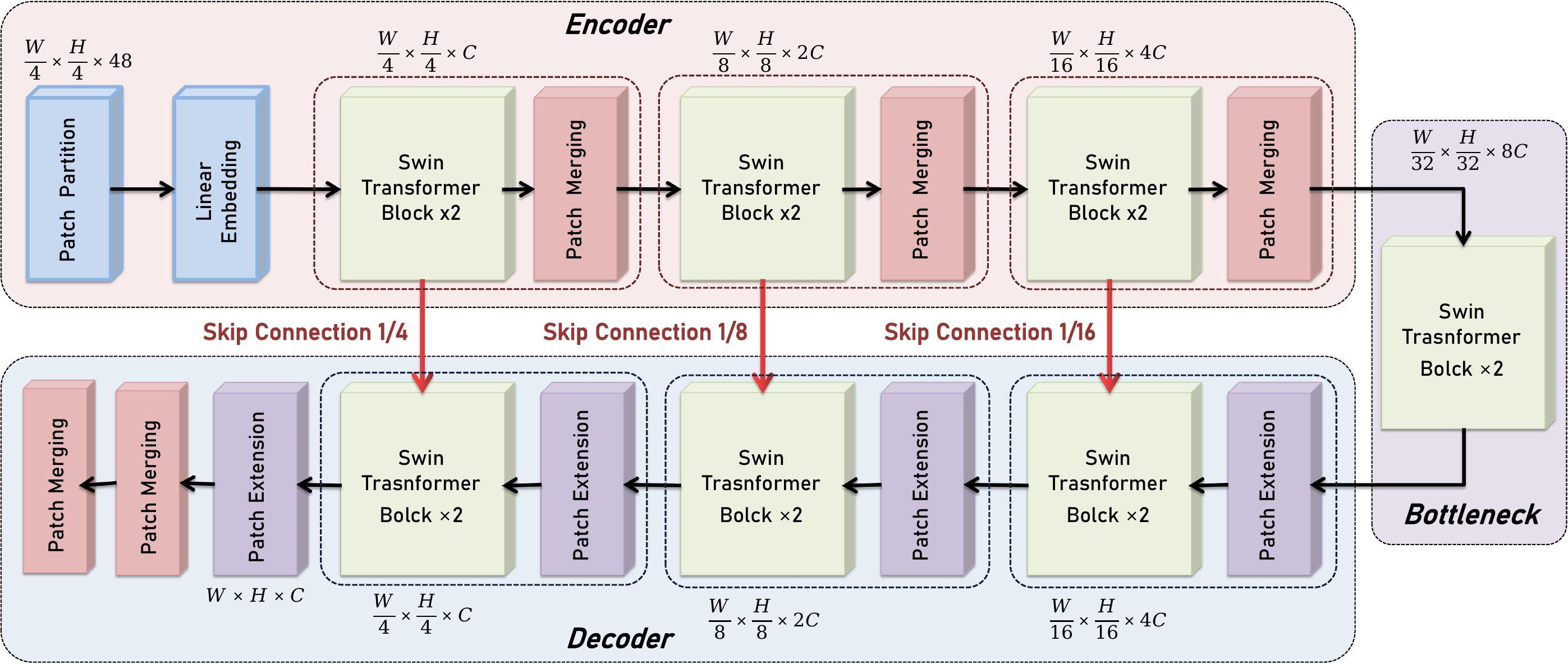}
	\caption{The architecture of \emph{Swin-Descriptors}, which is composed of encoder, bottleneck, decoder and skip connections. Encoder, bottleneck and decoder are all constructed based on swin transformer block.}
	\label{fig3}
\end{figure*}

The first step of Local Match is to apply a dense feature extractor $ \mathcal S$ on candidate images or input query image  $\emph{I}$ to obtain a 3D tensor $ \emph{D} = \mathcal{D}(\emph{I}), \emph{D} \in \mathbb{R}^{h \times w \times n}$, where $h \times w$ is the spatial resolution of the feature maps and n the number of channels. Our dense feature extractor architecture is presented in Fig. \ref{fig3}, which leverage the advantage of the shift window attention in transformer and skip connection.

% encoder
\textbf{Encoder:} In the encoder, the tokenized inputs with a dimensionality of $C$ and a resolution of $ \frac{W}{4} \times \frac{H}{4} $ are inputted into two consecutive Swin Transformer blocks for representation learning. Simultaneously, the Patch Merging layer reduces the number of tokens (downsampling by a factor of 2) and increases the feature dimension to twice the original dimension. This entire procedure is repeated three times in the encoder.

In this step, Patch Merging layer makes input patches are partitioned into four segments and combined. This process downsamples the feature resolution by a factor of 2. Additionally, since the concatenation operation leads to a fourfold increase in feature dimension, a linear layer is employed on the concatenated features to unify the feature dimension to twice the original dimension.

% Bottleneck
\textbf{Bottleneck:} To avoid convergence issues caused by excessive depth of transformer, the model employs only two consecutive Swin Transformer blocks to form a bottleneck to extract deep feature. Within this bottleneck, the feature dimension and resolution remain constant.

% decoder
\textbf{Decoder:} Corresponding to the encoder, use the Swin Transformer block to build a symmetric decoder. Unlike the Patch Merging layer used in the encoder, the decoder utilizes patch expansion layers to upsample the extracted depth features. The patch expansion layer reshapes feature maps of adjacent dimensions into higher-resolution feature maps (upsampled by a factor of 2) while reducing the feature dimensions to half of the original dimensions.

Regarding Patch Extension, take the first patch extension layer as an example. Before upsampling, the input feature with dimension $( \frac{W}{32} \times \frac{H}{32} \times 8C) $ undergoes a linear layer operation to increase the feature dimension to 2 times the original size, The resulting size is $( \frac{W}{32} \times \frac{H}{32} \times 16C) $. Subsequently, a rearrangement operation is applied to expand the resolution of the input features to 2 times the input resolution, while reducing the feature dimension to a quarter of the input dimension, yielding $( \frac{W}{16} \times \frac{H}{16} \times 4C)$.

% Skip Connection
\textbf{Skip connection:} Similar to many dense prediction vision tasks \cite{unet,densedepth2018high}, skip connections are employed to combine the multi-scale features of the encoder with the upsampled features. Shallow features and deep features are concatenated to alleviate the loss of spatial information caused by downsampling. Subsequently, a linear layer is applied to keep the dimensions of the connected features the same as those of the upsampled features.

\begin{figure}[!t]
	\centering
	\includegraphics[width=2in]{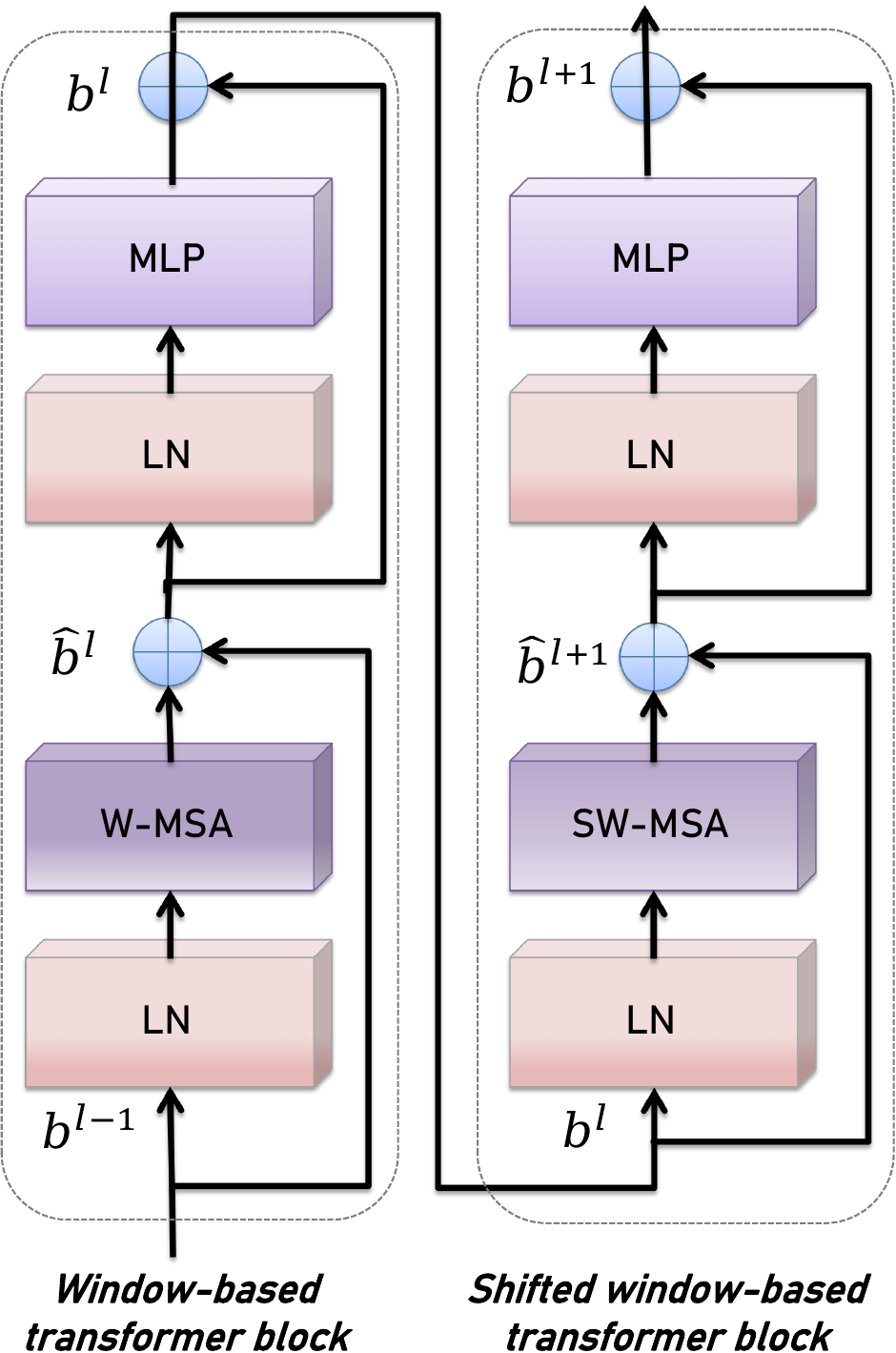}
	\caption{Swin transformer block}
	\label{fig4}
\end{figure}

% Swin Transformer Block
\textbf{Swin transformer block:} Differing from the conventional multi-head self-attention (MSA) module, the swin transformer block is structured around shifted windows. In Fig. \ref{fig4}, we illustrate two consecutive Swin Transformer blocks. Each swin transformer block comprises a LayerNorm (LN) layer, a multi-head self-attention module, a residual connection, and a 2-layer MLP with GELU non-linearity. In the two successive transformer blocks, the shifted window-based multi-head self-attention (W-MSA) module and the shifted window-based multi-head self attention (SW-MSA) module are applied, respectively. Leveraging this window partitioning mechanism, continuous swin transformer blocks can be formulated as follows:

\begin{equation}
\label{swin transformer block-1}
{\hat{b}}^{l} = W - MSA(LN(b^{l-1}))+b^{l-1},
\end{equation}

\begin{equation}
\label{swin transformer block-2}
b^{l} = MLP(LN({\hat{b}}^{l}))+{\hat{b}}^{l},
\end{equation}

\begin{equation}
\label{swin transformer block-3}
{\hat{b}}^{l+1} = SW -MSA(LN(b^{l}))+b^{l},
\end{equation}

\begin{equation}
\label{swin transformer block -4}
b^{l+1} = MLP(LN({\hat{b}}^{l+1})) + {\hat{b}}^{l+1},
\end{equation}
where $ {\hat{b}}^{l} $ and $b^{l}$ represent the outputs of the (S)W-MSA module and the MLP module of the lth block, respectively. 

\subsubsection{Keypoints Detection and Description}

The 3D dense feature output by \emph{Dense Feature Extraction} is simultaneously utilized for both keypoint detection and description. To enhance their accuracy and robustness, we introduce soft detection score under global semantic attention in this step.
%
% Unfortunately, both of these challenges happen to be present in our visual localization in remote sensing. Additionally, during the execution of UAVs flight missions, scale variations usually occur, which traditional transformers may not effectively address because of the size of the patch is fixed.
%To enhance the robustness of features to scale variations and maintain concise,
Firstly, utilizing cyclic-shift window to achieve global contextual attention, the quantity of batched windows remains identical to that of regular window partitioning, ensuring efficiency, Swin Transformer\cite{swin_transformer_2021} demonstrated it.
The windows are organized in a way that divides the image evenly without overlapping. Assuming that each window contains $P \times P$ patches, the computational complexity of the global multi-head self-attention (MSA) module and the window-based model on an image consisting of $h \times w$ patches is as follows:
\begin{equation}
	\label{w-msa}
	\begin{aligned}
	  &\text{MSA: }O(4hwC^2 + 2(hw)^2C), \\
	  &\text{W-MSA: }O(4hwC^2 + 2M^2hwC),
	\end{aligned}
\end{equation}
the former exhibits quadratic complexity with respect to the patch count hw, while the latter maintains linearity when the fixed value of $P$ is used (defaulted to 7). Global self-attention computations are typically impractical for a large $hw$, whereas window-based self-attention is scalable.

Similar to the previous works \cite{yuan2021tokens,liu2022swin}, a relative position bias matrix is adopted in each head computing similarity:

\begin{equation}
\label{swin attention}
Attention(Q,K,V) = SoftMax(\frac{QK^{T}}{\sqrt{d}}+\mathcal{A})V, 
\end{equation}
where $Q, K, V \in {\mathbb{R}}^{W^{2}\times d} $ denote the query, key and value matrices. $W^{2}$ and $d$ represent the number of patches in a window and the dimension of the $query$ or $key$, respectively. And, the values in $\mathcal{A}$ are taken from the bias matrix $\hat{\mathcal{A}} \in {\mathbb{R}}^{(2B-1)\times(2B+1)}$ instead of traditional $\mathcal{A} \in {\mathbb{R}}^{(B^2 \times B^2}$, because of the relative position along each axis lies in the range $[-B + 1, B - 1]$.

% The the straightforward interpretation of the 3D tensor D is as a dense set of descriptor % vectors d:
% \begin{equation}
% \label{eq2}
% \mathbf{v}_{ij} =D_{ij}, \mathbf{v} \in \mathbb{R}^n,
% \end{equation}
% with $i =1,...,h$ and $j = 1,...w$. These descriptor vectors can be easily compared between query and candidate images by calculating their Euclidean distances, allowing us to establish keypoints' correspondences. During training, these descriptors are fine-tuned to ensure that corresponding points in the scene produce similar descriptors, even in the presence of significant appearance variations. When inference, we apply an L2 normalization on the descriptors prior to comparing them:$\tilde{\mathbf{v}}_{ij}=\mathbf{v}_{ij}/ \lVert \mathbf{v}_{ij}\rVert_2$.
%

Furthermore, the above feature extraction function $ \mathcal{F} $ can be conceptualized as comprising $n$ different feature detection functions  $\mathcal{R}_k$, each producing a 2D response map $R_k$. These detection response maps are analogous to the Difference-of-Gaussians (DoG) responses found in SIFT. We post-process these raw scores through our soft detection score to select only a subset of locations as the output keypoints. This process is detailed below.
Contrary to traditional feature detectors that reduce detection map through a straightforward non-local-maximum suppression, in our case, there are multiple detection maps $R_k (k = 1, . . . , n)$, and a detection can occur on any of them. Therefore, for a point $p_{(i, j)}$ to be detected, we require:
\begin{equation}
	\label{eq4}
	\begin{aligned}
    {R_{ij}}^k \mbox{ is a local max in } R^k,  
	\mbox{with } k = \mathop{\arg\max}\limits_{t} {R_{ij}}^t .
	\end{aligned}	
\end{equation}
Obviously, for each pixel $(i, j)$, first selecting the most preeminent detector $\mathcal{R}^k$ (channel selection), and then verifying whether a local-maximum exists at position $(i, j)$ on that specific detector's response map ${R}^k$.

In order to take both criteria into account, we maximize the product of both scores across all feature maps $k$ to obtain a single score map. Furthermore, we enhance this detection strategy to make it compatible with neural network backpropagation. As illustrated in following formula:

\begin{equation}
\label{eq7}
	\left\{
	\begin{aligned}
		\phi_{ij} &= \max\limits_{k}(\eta_{ij}^{k}\zeta_{ij}^{k}) , \\
		\eta_{ij}^{k} &= \frac{exp(R_{ij}^{k})}{\sum_{(i^{'},j^{'})\in\mathcal{M}(i,j)}exp(R_{i^{'}j^{'}}^{k})}, \\
		\zeta_{ij}^{k} &= \frac{R_{ij}^{k}}{\max\limits_{t}R_{ij}^{t}},
	\end{aligned}
	\right.
\end{equation}
where $\phi_{ij}$ represents a soft-local-max, $ \mathcal{M}(i,j) $ is the set of $9$-pixel neighborhood of $(i, j)$, including the pixel itself. And $\zeta_{ij}^{k}$ computes a ratio-to-max for soft channel selection. 

Finally, the soft detection score $s_{ij}$ at a pixel $(i, j)$ with global semantic attention is obtained by performing an image-level normalization:
\begin{equation}
	\label{eq8}
	s_{(i,j)} = \frac{\phi_{ij}}{\sum\limits_{i{'},j{'}}\phi_{i{'}j{'}}}.
\end{equation}

\subsubsection{Jointly Optimizing Detection and Description}

This section describes the loss for detection and description optimizing. In detection, we aim for keypoint repeatability across changes in viewpoint and illumination, while in description, we seek distinctive descriptors to prevent mismatches. In order to meet these requirements, we propose an extend triplet edge ranking loss with pixel geometric attention. We will begin by revisiting the triplet edge ranking loss and then introduce our extended version that combines both detection and description aspects.

For a pair of query and reference images, $(I_{q}$ and $I_{r}^{1} )$, and a correspondence $c \colon X\leftrightarrow Y $ between them, where $X \in I_{q}, Y \in I_{r}^{1}$, our extend triplet margin ranking loss aims to minimize the distance between corresponding descriptors ${\hat{R}}_{X}^{(1)}$ and ${\hat{R}}_{Y}^{(2)}$, while maximizing the distance to other confusing descriptors ${\hat{R}}_{N_{1}}^{(1)}$ or ${\hat{R}}_{N_{2}}^{(2)}$ in either image, which might exist due to presence of similar-looking image structures. Towards this objective, we define the positive descriptor distance $d_{p(c)}$ between the corresponding descriptors as:
\begin{equation}
	\label{loss-1}
	d_{p(c)} = {\lVert {\hat{R}}_{A}^{(1)} - {\hat{R}}_{B}^{(2)} \rVert}_{2}.
\end{equation}

The negative distance $d_{n(c)}$, which accounts for the most confounding descriptor for either ${\hat{R}}_{X}^{(1)}$ or ${\hat{R}}_{Y}^{(2)}$, is defined as:

\begin{equation}
	\label{loss-2}
	d_{n(c)} = \min{\Bigl( {\lVert {\hat{R}}_{N_{1}}^{(1)} - {\hat{R}}_{N_{2}}^{(2)} \rVert}_{2}, {\lVert {\hat{R}}_{X}^{(1)} - {\hat{R}}_{Y}^{(2)} \rVert}_{2} \Bigr)},
\end{equation}
where the negative samples ${\hat{R}}_{N_{1}}^{(1)}$ and ${\hat{R}}_{N_{2}}^{(2)}$  are the hardest negatives that lie outside of a square local neighbourhood of the correct correspondence:
\begin{equation}
	\label{loss-3}
	N_{1} = \mathop{\arg\min}\limits_{P \in I_{q}}{\lVert {\hat{R}}_{P}^{(1)} - {\hat{R}}_{B}^{(2)} \rVert}_{2} s.t. \lVert P-X \rVert_{\infty} \textgreater K ,
\end{equation}
and similarly for $N_2$. The triplet margin ranking loss $\mathcal{L_M}$ for a margin $\mathcal{M}$ can be defined as:

\begin{equation}
	\label{loss-4}
	\mathcal{L_M}(c) = \max (0, margin+ d_{p(c)}^{2}- d_{n(c)}^{2}).
\end{equation}

% Intuitively, the triplet margin ranking loss promotes descriptor distinctiveness by penalizing confounding descriptors that could result in incorrect match assignments.
To further encourage detection repeatability against appearance variations, a soft detection term is incorporated into the triplet margin ranking loss as follows:

\begin{equation}
	\label{loss-5}
	\mathcal{L}(I_{q}, I_{r}) = \sum \limits_{c \in \mathcal{C}} \frac{s_{c}^{(1)}s_{c}^{(2)}}{\sum_{q \in \mathcal{C}}s_{q}^{(1)}s_{q}^{(2)}} \mathcal{L_M}(p(c),n(c)),
\end{equation}
where $s_{c}^{(1)}$ and $s_{c}^{(2)}$ are the soft detection scores \ref{eq8} at points $X$ and $Y$ in $I_{q}$ and $I_{r}^1$, respectively, and $\mathcal{C}$ is the set of all correspondences between $I_{q}$ and $I_{r}^1$. 

The proposed loss calculates a weighted average of $\mathcal{M}$ for all matches, determined by their soft detection scores. Consequently, to minimize the loss, correspondences that exhibit higher distinctiveness (indicated by a lower margin term) will receive elevated relative scores, and conversely, correspondences with higher relative scores are incentivized to possess descriptors that stand out from the rest.

\subsection{Reranking and Matching}

In stereo matching, using RANdom SAmple Consensus (RANSAC) or similar geometric constraint algorithms is an effective method for refining matching points. However, when the set of matching point pairs to be refined contains a large number of significantly mismatched points, the RANSAC algorithm's random sampling and iterative approach can become highly unstable due to the influence of these erroneous matches. Therefore, prior to applying geometric constraints, it is necessary to perform coarse refinement of the matching point pairs. 

This section will introduce the details of our stereo matching and reranking strategy based on the keypoints and descriptors output by the above \emph{Swin-Descriptors}.

\subsubsection{Adaptive Distance Filter}
Through the Fast Library for Approximate Nearest Neighbors (FLANN) algorithm, we search for the nearest neighboring matching pair, which includes the closest first matching point and the second closest second matching point, from the set of matching pairs to be filtered. Typically, for the $j$-th matching pair in the set to be filtered, a smaller Euclidean distance ($dis_j$) between the first matching point and a larger Euclidean distance ($dis_{j}^{'}$) between the second matching point implies better match quality. Traditional algorithms often employ a fixed ratio factor $t$ as a threshold, i.e., when $dis_{j} < t·dis_{j}^{'}$, the pair is selected as a candidate matching pair. However, due to substantial differences in image quality arising from different sensors and temporal factors, it is challenging to anticipate the range of Euclidean distance differences when searching for Euclidean space distances with deep learning features.

As a result, for each pair of images, it is typically necessary to manually adjust the $t$ threshold value repeatedly to achieve suitable values for filtering high-quality matching pairs. To address this issue and improve the algorithm's adaptability, this paper introduces a dynamic and adaptive Euclidean distance constraint method. This method statistically analyzes the data of the matching pairs to be refined and automatically configures corresponding parameters based on data characteristics.

First, from the $N$ matching pairs retrieved by the FLANN search, which may include a significant number of mismatches, we compute the mean difference between the distances of the first matching point and the second matching point.

\begin{equation}
	\label{dynamic filter}
	dis^* = (\sum_{j=1}^{N} dis_{j}^{'}-dis_{j})/N.
\end{equation}

For each matching pair to be filtered, the removal condition is based on the difference between the first distance ($dis_{j}$) being smaller than the second distance ($dis_{j}^{'}$) and the mean difference ($dis^*$) in distances. The formula is as follows:

\begin{equation}
	\label{filter-2}
	dis_{j} \textless dis_{j}^{'} -dis^*.
\end{equation}

The algorithm utilizes the mean difference in distances calculated from the data as a discriminative comparison criterion. This approach adapts well to variations between image pairs from different domains, enabling effective initial filtering to retain high-quality matching points. At the same time, enhancing the stability of the RANSAC output.

\subsubsection{RANSAC via Affine Model}
In RANSAC, the selection of geometric constraint relationships should be based on the imaging geometry of the images being matched. In practical engineering applications, it is advisable to use rigorous constraint models whenever possible. For instance, when dealing with images captured from Area-Array Cameras, models like homography matrices or essential matrices are often suitable constraints. On the other hand, for satellite images captured from Line-Array Cameras, constraints like the RPC (Rational Polynomial Coefficients) model or polynomial models based on kernel lines are commonly used.

In our experiments, given the relatively large photographic distance of the selected remote sensing images, minimal ground height differences, and a small image region, we chose to utilize the affine transformation model. This choice allows for accommodating various types of transformations, including scaling, translation, rotation, and shearing, between image pairs originating from different imaging models.

\subsubsection{Reranking}

After obtain the distinctive keypoints and associated stereo matching relationships, we calculate the average pixel distance of final correspondences between the query image and the candidate images. Subsequently, we return the top 5 reference images with the smallest average distance as the refined retrieval candidates.

\subsection{Pose Estimation}
After obtaining stereo matching relationships between query and candidate images, the precise geolocation of UAV can be calculated by solving a PnP problem, as illustrated in Fig. \ref{outline}(c). 

First of all, given the geo-tags and the keypoints of the five satellite images, it is not difficult to obtain the GPS coordinates of these keypoints, that is, the GPS coordinates of corresponding keypoints in the query image. And then, by solving PnP problem of the 2D and 3D correspondences of keypoints in the query image, the camera pose of UAV is estimated.

By leveraging the PnP principle, we can establish the motion model of the camera's coordinate system relative to the world coordinate system, which corresponds to the camera pose. And this camera pose represents the drone's pose, the translation component of this pose provides the precise geolocation of this drone.

To find the optimal camera pose, we formulate a nonlinear least squares problem based on the re-projection error. This problem aims to minimize the sum of errors between the $n$ matching points re-projected pixel coordinates in the image plane and their corresponding projected pixel coordinates computed using the camera pose and the corresponding 3D point positions. The re-projection error is defined as follows:
\begin{equation}
	\label{pnp}
	\boldsymbol{T^{*}} = \mathop{\arg\min}\limits_{\boldsymbol{T}} \frac{1}{2} \sum_{i=1}^{n}{\left\lVert \boldsymbol{p}^{i}-\frac{1}{s_{i}}\boldsymbol{KT}\boldsymbol{P}_{i}^{W}\right\rVert}_{2}^{2}.
\end{equation}
In the equation, $\boldsymbol{K}$ and $\boldsymbol{T}$ respectively represents the camera intrinsic matrix and transformer matrix, $\boldsymbol{p}^{i}$ represents the pixel coordinates of a point, $ \boldsymbol{P}_{i}^{W} $ represents the corresponding world coordinates of a matching point, and $s_i$ is the depth of $\boldsymbol{p}^{i}$ in camera coordinate system. These two sets of coordinates represent different representations of the same spatial point in different coordinate systems. $ \frac{1}{s_{i}}\boldsymbol{KT}\boldsymbol{P}_{i}^{W} $ is the coordinates of the point in the world coordinate system, and is the projected coordinates in the pixel coordinate system, obtained through the camera's motion pose.

By minimizing the re-projection error, we can obtain an estimate of the optimal camera pose $\boldsymbol{T^{*}}$, that is, the ultimate precise geolocation.
\section{Experiments}

In this section, we run extensive experiments to show the robustness and accuracy of the proposed \emph{CurriculumLoc} compared to existing state-of-the-art techniques by evaluating on two challenging datasets. In what follows, we present implementation details, datasets, evaluation metrics, performance comparisons and ablation studies.

\subsection{Implementation Details}

We implemented our approach in Pytorch and resize all images to $500 \times 500$ pixels.

\subsubsection{Training Details} 

During training, the last layer of \emph{Swin-Descriptors} was fine-tuned for 50 epochs. We employed the Adam optimizer with an initial learning rate of $10^{-3}$, which was subsequently halved every 10 epochs. For each pair, we selected a random $224 \times 224$ crop centered around a correspondence point, and the patch size was set to 4. We utilized a batch size of 1 and ensured that the training pairs contained a minimum of 128 correspondences to yield meaningful gradients.

\subsubsection{Test Details} 

At test time, to increase the resolution of the feature maps, we remove the last patch merging layer. This adjustment result in feature maps with one-fourth the resolution of the input, enabling the detection of more potential keypoints and improving localization. The positions of the detected keypoints were refined locally at the feature map level, following a methodology akin to that employed in SIFT. Finally, the descriptors were bilinearly interpolated at the refined positions.

\subsubsection{Datasets} 

We evaluated our approach on one public benchmark dataset: ALTO and our self-collected dataset: TerraTrack. All of these datasets contain various challenging environmental variations. Table \ref{tab:dataset} summarizes the qualitative nature of them. All images are resized to 500 × 500 while evaluation. 

\begin{table}[!t]
	\caption{\textbf{Summary of datasets used for training and evaluation.} + indicates that the dataset contains the particular environmental variation, and - is the opposite.\label{tab:dataset}}
	\centering
	\begin{tabular}{llllllllll}
		\toprule
		\multicolumn{2}{c}{Dataset} & \multicolumn{3}{c}{Environment} & \multicolumn{5}{c}{Variation} \\
		\cmidrule(lr){3-5} 	\cmidrule(lr){6-10}
		& & \rotatebox{90}{Urban} & \rotatebox{90}{Suburban} & \rotatebox{90}{Natural}  & \rotatebox{90}{Viewpoint} & \rotatebox{90}{Illumination} & \rotatebox{90}{Scale} & \rotatebox{90}{Texture} & {\rotatebox{90}{Domain}} \\
		\midrule
		\multicolumn{2}{c}{ALTO \cite{ALTO_Scherer_2022} }   &  & \checkmark & \checkmark & + &  + & - & + & +  \\
		\midrule
		\multirow{16}{*}{\rotatebox{90}{TerraTrack}}                  & mavic-river               &    & \checkmark & \checkmark & + &  + & - & - & +  \\
		& mavic-hongkong            & \checkmark &   &    & + &  + & - & - & +  \\
	    & mavic-factory             &    & \checkmark &   & + &  + & - & - & +  \\
		& mavic-fengniao            &   & \checkmark &    & + &  + & - & - & +  \\
		& inspire1-rail-kfs         &   & \checkmark & \checkmark & + &  + & - & + & +  \\
		& phantom3-grass-kfs        &   & \checkmark & \checkmark & + &  + & - & + & +  \\
		& phantom3-centralPark-kfs  &    & \checkmark & \checkmark & + &  + & - & - & +  \\
		& phantom3-npu-kfs          &   & \checkmark &  & + &  + & + & - & +  \\
		& phantom3-freeway-kfs      &    & \checkmark & \checkmark & + &  + & - & + & +  \\
		& phantom3-village-kfs      &    & \checkmark & \checkmark & + &  + & - & - & +  \\
		& gopro-npu-kfs             &   & \checkmark &   & + &  + & + & - & +  \\
		& gopro-saplings-kfs        &   & \checkmark &   & + &  + & - & + & +  \\
		& mavic-xjtu                & \checkmark &    &   & + &  + & - & - & +  \\
		& phantom3-huangqi-kfs      &   & \checkmark & \checkmark & + &  + & - & + & +  \\
		& mavic-npu                 &   & \checkmark &   & + &  + & + & - & +  \\
		& mavic-road                &   & \checkmark &    & + &  + & - & - & +  \\
		\bottomrule
	\end{tabular}
	\label{tabel_datasets}
\end{table}

For our \emph{TerraTrack}, a self-made hexacopter and a DJI Phantom3 are used to record the query images in different terrains and heights, and the reference images derive from the satellite imagery of \emph{Google Maps}. 
The current dataset is available at the link given in the Abstract, and can serve for cross-domain geolocalization algorithm validation. However, it's important to note that the dataset, while valuable, is not extensive, and the full dataset is forthcoming.

In order to generate training data for pixel-wise correspondences, we utilized our TerraTrack dataset, which comprises 16 distinct scenes reconstructed from a total of 5,848 UAV images and 23,392 satellite images employing COLMAP \cite{colmap1_schonberger2016pixelwise, colmap2_schonberger2016structure}. 

To extract correspondences, we initially evaluated all pairs of images exhibiting a minimum of 50\% overlap in the sparse SfM point cloud. For each pair, we projected all depth-informed points from the second image onto the first image. A depth-check procedure, relative to the depth map of the first image, was executed to eliminate occluded pixels. This dataset was subsequently divided into a validation dataset, consisting of 359 image pairs (from two scenes, each containing fewer than 100 image pairs), and a training dataset comprising the remaining 14 scenes.

\subsection{Comparison with Other Methods}

\subsubsection{Compared Methods}
We compare our rerank results with several state-of-the-art vision place recognition algorithms, including retrieval with global descriptors: NetVLAD \cite{Arandjelovic_Gronat_Torii_Pajdla_Sivic_2018} and SFRS \cite{SFRS} and two-stage pipeline (global retrieval and rerank based local descriptors): Patch-NetVLAD \cite{PatchNetVLAD}, DELG \cite{delg_2020} and TransVLAD \cite{xu2023transvlad}. For Patch-NetVLAD, we evaluated both 2048 and 4096 descriptor dimension, denoted as Patch-NetVLAD-2048 and Patch-NetVLAD-4096 respectively. Moreover, we also compared against a latest global-based method GCG-Net \cite{zhang2023gcg} and a two stage pipeline TransVLAD \cite{xu2023transvlad}. GCG-Net \cite{zhang2023gcg} employed graph structure to represent global descriptors and TransVLAD  \cite{xu2023transvlad} extracted feature maps by a transformer based network. For all methods, we use recall@N metric which computes the percentage of query UAV images that are correctly localized. A query image is considered to have been correctly positioned if at least one of the first $N$ reference images is within a threshold distance from the query's ground truth location. 
\subsubsection{Metrics}
For the ALTO and TerraTrack datasets, we employ the recall@N metric, which calculates the percentage of query images that achieve accurate localization. A query image is considered as correctly localized if at least one of the top N ranked reference images is within a threshold distance from the ground truth location of the query. We respectively set the threshold distance at 20 and 50 meters, based on the flight altitude and range specifications in our datasets. The setting of these two distance thresholds provides a more comprehensive evaluation of the algorithm's robustness. 
\subsubsection{Results}
\begin{figure}[!t]
	\centering
	\includegraphics[width=3.in]{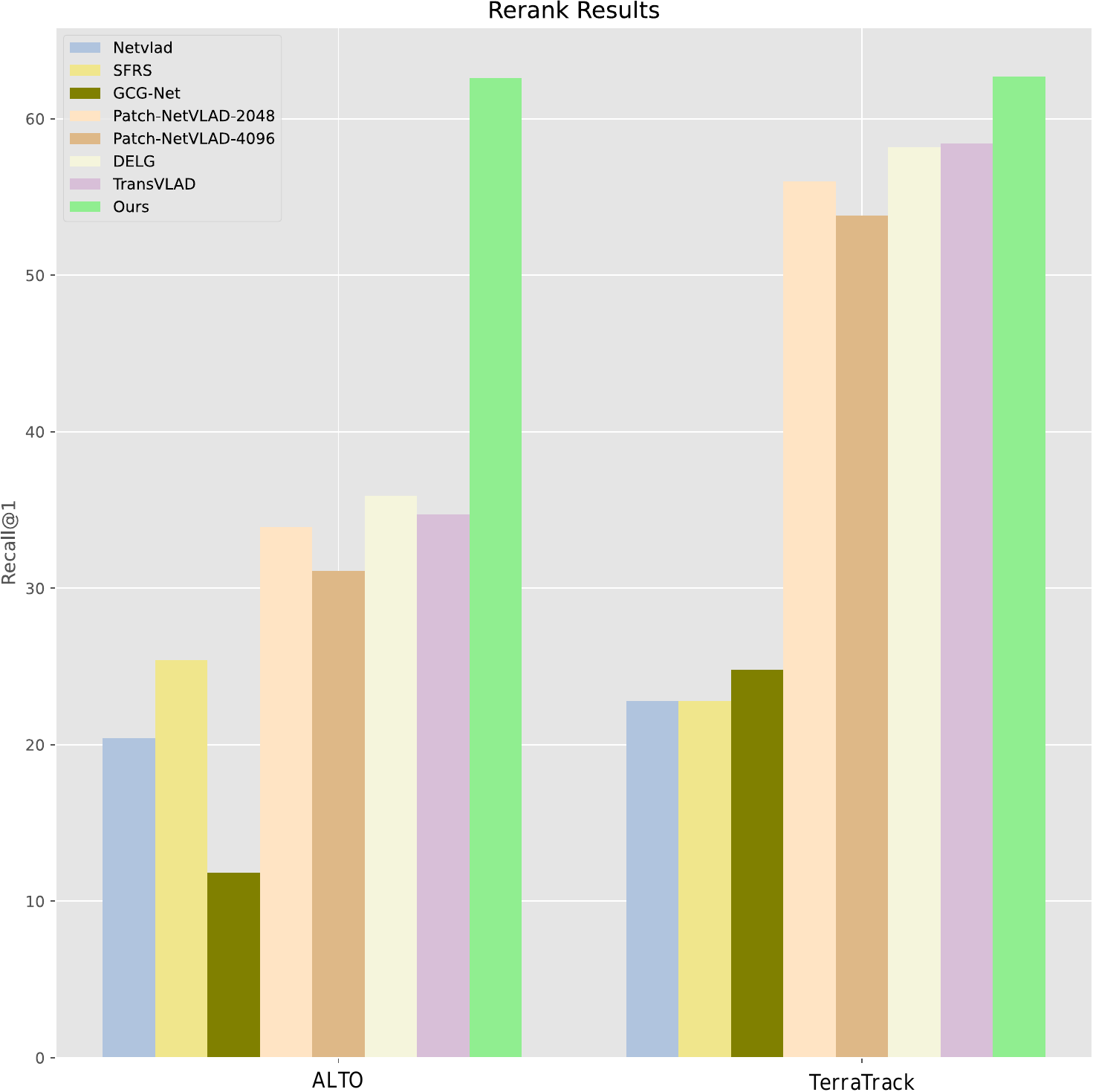}
	\caption{Comparison results of retrieval recall@1 with dist=20$m$.}
	\label{fig:dist20_bar}
\end{figure}

\begin{figure}[!t]
	\centering
	\includegraphics[width=3.0in]{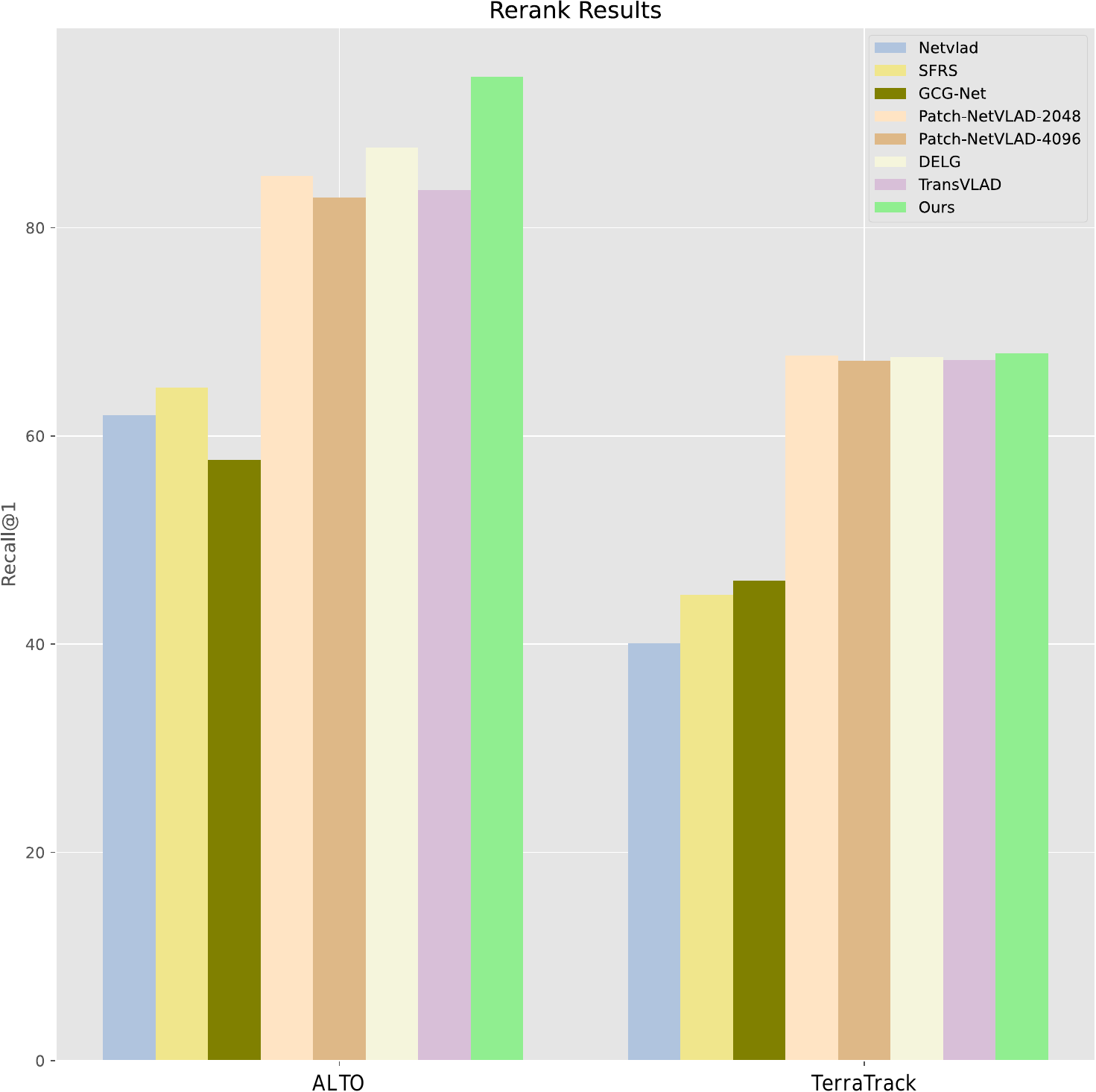}
	\caption{Comparison results of retrieval recall@1 with dist=50$m$.}
	\label{fig:dist50_bar}
\end{figure}

\begin{figure*}[!t]
	\centering
	\includegraphics[width=7.0in]{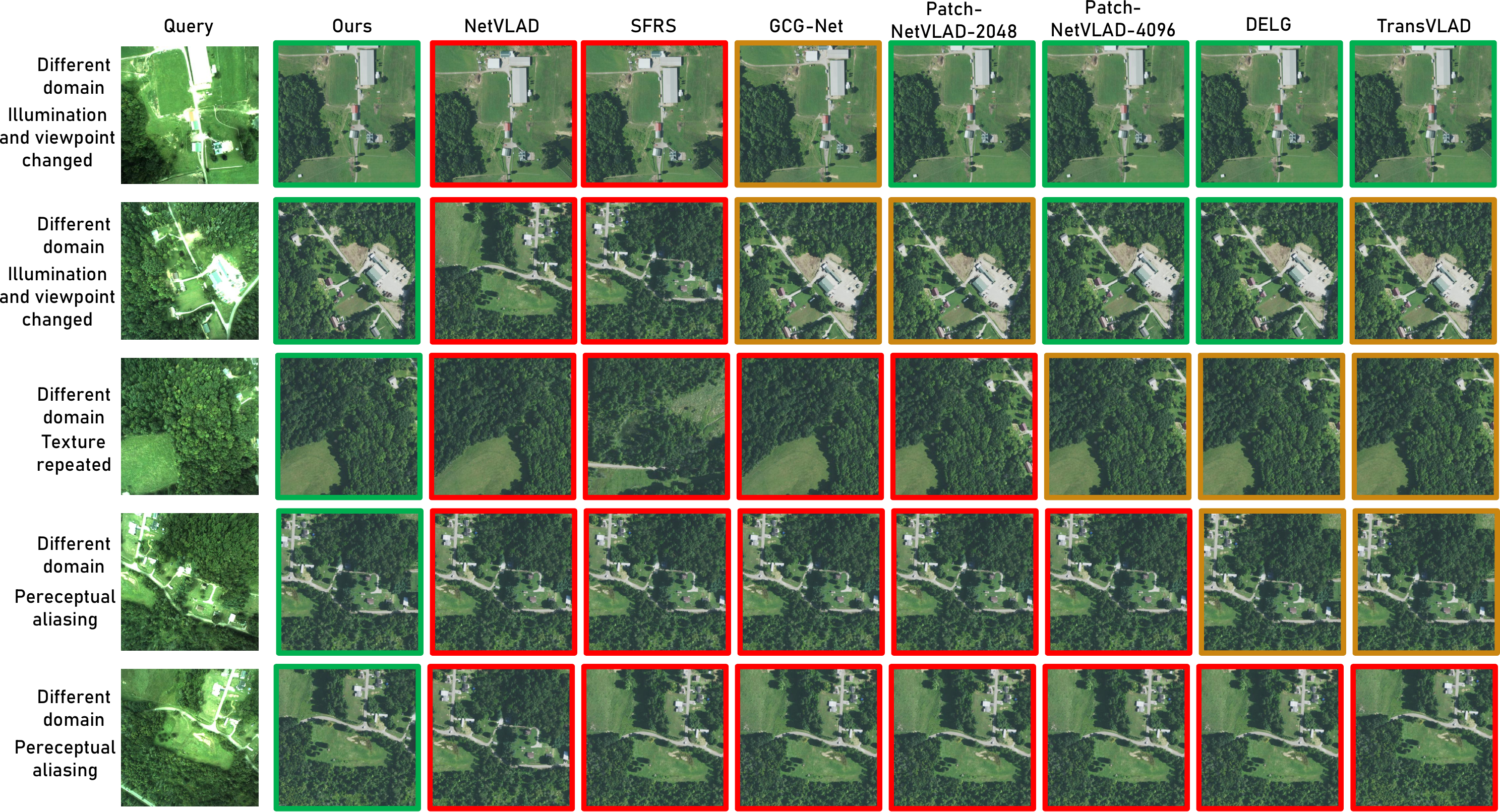}
	\caption{\textbf{Comparison of retrieval results on ALTO validation dataset.} In these challenging examples, Our CurriculumLoc successfully retrieves the matching database image, while all other methods produce false results.}
	\label{fig:alto_pr}
\end{figure*}

\begin{figure*}[!t]
	\centering
	\includegraphics[width=7.0in]{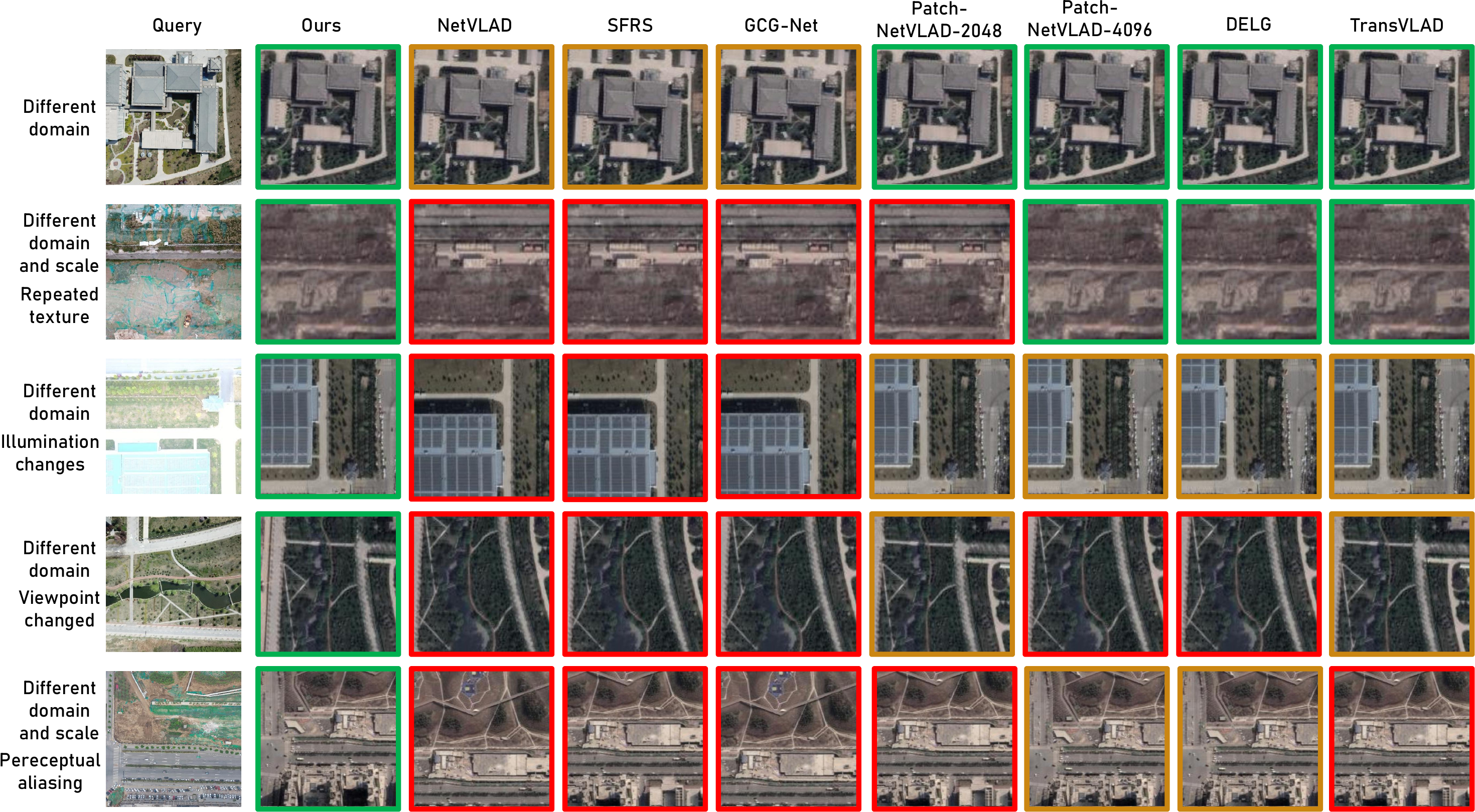}
	\caption{\textbf{Comparison of retrieval results on TerraTrack validation dataset.} In these challenging examples, Our CurriculumLoc successfully retrieves the matching database image, while all other methods produce false results.}
	\label{fig:terratrack_pr}
\end{figure*}

The quantitative results of ALTO and TerraTrack are shown in Table \ref{tab:table-vpr-20} and Table \ref{tab:table-vpr-50}. In addition, Fig. \ref{fig:dist20_bar} and Fig. \ref{fig:dist50_bar} show the recall@1 preformance on ALTO and TerraTrack. In conclusion, compare against other retrieval-based methods our two stage pipeline achieves a significant improvement with 62.6\%/94.5\% of R@1 and 92.6\%/99.7\% of R@5 in ALTO with metric distance of $20m/50m$, and 62.7\%/67.9 \% of R@1 in TerraTrack with metric distance of $20m/50m$, respectively. Fig. \ref{fig:alto_pr} and Fig. \ref{fig:terratrack_pr} show the the top-1 retrieval reference images of query images by our model with NetVLAD, SFRS, GCG-Net, Patch-NetVLAD-2048, Patch-NetVLAD-4096, DELG and TransVLAD on challenging scenes, besides different domian (query images from UAV capture, reference images is a selection of satellite images), repeated texture, illumination and viewpoint changed, and pereceptual aliasing).

\begin{table*}[!t]
	\caption{The results of local match with other place recognition methods (dist=$20m$)\label{tab:table-vpr-20}}
	\centering
	\begin{tabular}{llllll|llll}
		\toprule
		\multicolumn{2}{c}{Methods} & \multicolumn{4}{c}{ALTO} & \multicolumn{4}{c}{TerraTrack} \\
		\cmidrule(lr){3-10}
		&                     &R@1      & R@5     & R@10    & R@20 &R@1      & R@5     & R@10    & R@20   \\
		\midrule
		\multirow{3}{*}{{Global}}                                                          & NetVLAD \cite{Arandjelovic_Gronat_Torii_Pajdla_Sivic_2018}            & 20.4 & 58.3 & 77.1 & 88.4 & 22.8& 48.8& 57.8& 67.8 \\
		& SFRS \cite{SFRS}           & 25.4 & 60.1 & 78.6 & 91.4 & 22.9 &49.1 & 58.1& 68.9 \\
		& GCG-Net \cite{zhang2023gcg}            & 11.8 & 36.9 & 51.9 & 66.0 & 24.8 &49.2 &56.1 &62.4 \\
		\midrule
		\multirow{4}{*}{{Local}}                                                          & Patch-NetVLAD-2048 \cite{PatchNetVLAD}      & 33.9 & 78.3 & 91.2 & 95.6 & 56.0& 68.6& 72.3& 74.4 \\
		& Patch-NetVLAD-4096 \cite{PatchNetVLAD}      &  31.1&74.8  &90.4  &94.9  & 53.8 & 64.7& 68.1 & 69.2\\
		& DELG \cite{delg_2020}      & 35.9 & 79.9 & 92.3 & 95.6 & 58.2&68.3 &71.9 &73.6 \\
		& TransVLAD \cite{xu2023transvlad}      & 34.7 & 78.8 & 91.9 & 95.3 & 58.4&67.9 & \textbf{72.1} & \textbf{73.8} \\
		\midrule
		\multicolumn{2}{c}{{Ours}}   & $\mathbf{62.6}$ & $\mathbf{92.6}$ & $\mathbf{95.6}$ & $\mathbf{96.1}$   & $\mathbf{62.7}$ & $\mathbf{69.2}$ & 70.2 & 72.7 \\
		
		\bottomrule
	\end{tabular}
\end{table*}

\begin{table*}[!t]
	\caption{The results of local match with other place recognition methods (dist=$50m$) \label{tab:table-vpr-50}}
    \centering
	\begin{tabular}{llllll|llll}
		\toprule
		\multicolumn{2}{c}{Methods} & \multicolumn{4}{c}{ALTO} & \multicolumn{4}{c}{TerraTrack} \\
		\cmidrule(lr){3-10}
		&                     &R@1      & R@5     & R@10    & R@20 &R@1      & R@5     & R@10    & R@20   \\
		\midrule
		\multirow{3}{*}{{Global}}                                                          & NetVLAD \cite{Arandjelovic_Gronat_Torii_Pajdla_Sivic_2018}            & 62.0 & 88.3 & 94.5 & 98.0 & 40.1& 61.5& 71.4& 78.9\\
		& SFRS \cite{SFRS}           & 64.6 & 90.6 & 93.4 & 98.0 & 44.7 &61.6 &72.5 &75.6\\
		& GCG-Net \cite{zhang2023gcg}            & 57.7 & 72.2 & 79.7 & 84.2 & 46.1 &62.1 &65.1&69.2\\
		\midrule
		\multirow{4}{*}{{Local}}                                                          & Patch-NetVLAD-2048 \cite{PatchNetVLAD}      & 85.0 & 97.4 & 98.8 & 99.8 & 67.7& 74.4& 78.6& 82.6\\
		& Patch-NetVLAD-4096 \cite{PatchNetVLAD}      & 82.9 & 96.9 & 98.2 & 99.0 & 67.2& 72.0&76.5 &79.1 \\
		& DELG \cite{delg_2020}      & 87.7 & 97.9 & 99.2 & 99.6 &67.6 & \textbf{75.1} &\textbf{78.6} &\textbf{82.4} \\
		& TransVLAD \cite{xu2023transvlad}      & 83.6 & 96.1 & 98.6 & 98.6 & 67.3 &73.5 &78.9 &81.9 \\
		\midrule
		\multicolumn{2}{c}{{Ours}} & $\mathbf{94.5}$ & $\mathbf{99.7}$ & $\mathbf{99.8}$ & $\mathbf{99.8}$   & $\mathbf{67.9}$ & 74.3 & 77.9 & 78.9    \\
		
		\bottomrule
	\end{tabular}
    \label{table-vpr-50}
\end{table*}

The above comparative experiments show that our method achieves state-of-the-art localization results in the rerank stage, and also demonstrates the effectiveness of our feature detection and extraction module \emph{Swin-Descriptors}. 

\subsection{Ablation Study}

\subsubsection{Initial Retrieval Candidates Number} 
\begin{figure}[!t]
	\centering
	\includegraphics[width=3.5in]{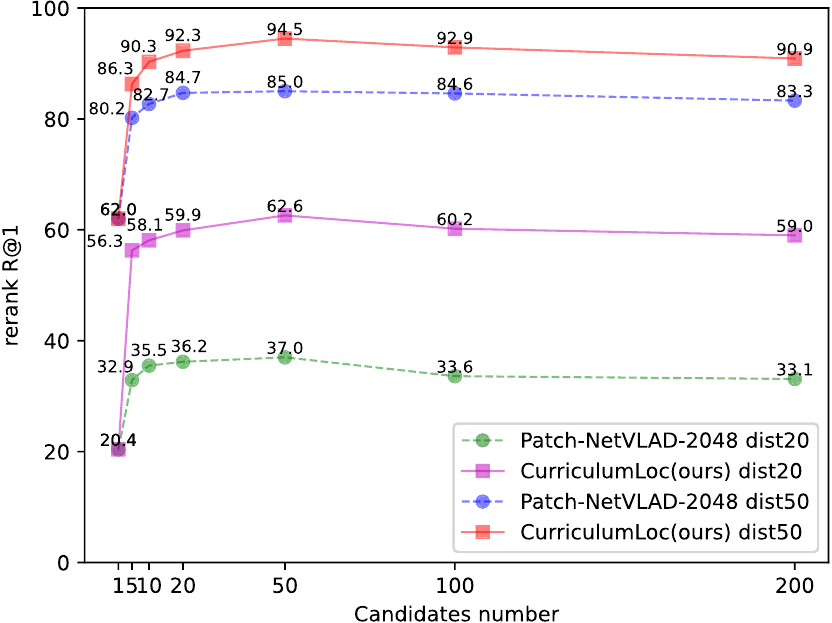}
	\caption{The rerank recall@1 preformance of our \emph{CurriculumLoc} and Patch-NetVLAD-2048 with different candidate numbers on ALTO.}
	\label{pr_res}
\end{figure}

Theoretically, the rerank efficiency is inversely proportional to the number of input candidate images, but the positioning accuracy of the rearrangement results is directly proportional to it. Fig. \ref{pr_res} shows the impact of different number of input candidates on the rerank result (recall@1) on the ALTO dataset. It is evident that $n=50$ is the optimal number of candidate images.

This experiment also shows that the initial retrieval in our pipeline can saving rerank efficiency of matching. Using top $50$ initial candidates to rerank can significantly improve the performance of recall@1 from 20.4\%/62.0\% to 62.6\%/94.5\% with distance threshold of $20m/50m$.

\subsubsection{Modules in Rerank and Match}
To verify the effectiveness of our \textit{Swin-Descriptors}, transform model in dynamic filter and rerank strategy, we conduct several ablation experiments to further validate the design and pipeline of our algorithm. The results on ALTO and TerraTrack datasets are reported in Table \ref{tab:ablation}.
\begin{table*}[!t]
	\caption{Ablations with our \textit{Swin-Descriptors}, dynamic filter strategy and project model on ALTO and TerraTrack datasets. \label{tab:ablation}}
	\centering
	\begin{tabular}{llllllllll}
		\toprule
		\multicolumn{2}{c}{Methods} & \multicolumn{4}{c}{dist20} & \multicolumn{4}{c}{dist50}\\
		\cmidrule(lr){3-10}
		& &R@1      & R@5     &R@10    & R@20 &R@1      & R@5     &R@10    & R@20  \\
		\midrule
		\multirow{8}{*}{ALTO} &
		VGG-16 + $F_{num}$ + ADF + affine	& 43.6 & 78.9 & 90.8 & 94.6 & 78.9& 90.3& 95.7& 98.0 \\
		& VGG-16 + $F_{dist}$ + ADF + affine	& 50.4 & 85.6 & 92.6 & 94.6 & 86.4&92.6 &96.8 &98.0 \\
		& SW + $F_{num}$ + ADF + affine	& 54.4 & 88.8 & 94.6 & 95.2 & 88.7& 95.6&97.8 &98.2 \\
		& SW + $F_{dist}$ + affine & 62.1 & 92.2 & 95.4 & \textbf{96.1} & 92.9 & 98.7 &99.5 &\textbf{99.8} \\
		& SW + $F_{dist}$ + ADF + euclidean  & 57.8 & 91.2 & 94.4 & 93.6 & 90.9&96.8 &98.8 & 99.2 \\
		& SW + $F_{dist}$ + ADF + similarity  & 61.8 & 92.6 & \textbf{95.6} & \textbf{96.1} &93.6 &99.6 &\textbf{99.8} & \textbf{99.8}\\
		& SW + $F_{dist}$ + ADF + project  & 62.3 & 92.5 & \textbf{95.6} & \textbf{96.1} & 94.1& 99.6&\textbf{99.8} &\textbf{99.8} \\
		& SW + $F_{dist}$ + ADF + affine  & \textbf{62.6} & \textbf{92.6} & \textbf{95.6} & \textbf{96.1} & \textbf{94.5} &\textbf{99.7} &\textbf{99.8} &\textbf{99.8} \\
		
		\bottomrule
	\end{tabular}
	\label{ablation}
\end{table*}
symmetric Encoder-Decoder Swin Transformer with shifted window attention mechanism and soft detection score strategy
Note that SW refers to the proposed symmetric transformer dense feature extraction network with shifted window attention in \textit{Swin-Descriptors}, while VGG-16 refers to extracting feature by VGG-16 without attention mechanism. $F_{num}$ and $F_{dist}$ refer to re-rank global retrieval results with the number of correspondences or the average pixel distance of all correspondences. ADF refers to our adaptive distance filter brfore RANSAC in \textit{dynamic filter}. Euclidean, similarity, project, affine are transform model of RANSAC in \textit{Geometric Verification}. 
From the results, we can draw the following conclusions. \textit{Swin-Descriptors} with SW can encode multi-scale level semantic feature and capture local information with global awareness and pixel supervision to improve the rerank performance. In particular, using SW to replace the VGG-16 improves the R@1 performance of our method from 50.4\%/78.9\% to 62.6\%/94.5\% on ALTO dataset with metric distance of $20m/50m$, respectively. Rerank the global candidates based on the average pixel distance of all correspondences ($F_{dist}$) is more robust than the number of correspondences($F_{num}$) between query image and candidate image. For example, exploiting the $F_{dist}$ improves the R@1 performance of the $F_{num}$ from 43.6\% to 50.4\% with VGG-16 and achieves 8.2\% improvement at R@1 with SW.
Adaptive distance filter (ADF) before RANSAC in \textit{Dynamic filter} improves the quality of matched keypoints between query and candidate image and thus improves the R@1 from 62.1\%(92.9\%) to 62.6\%(94.5\%) with metric distance of $20m/50m$.
Among these four transform models in RANSAC, the similarity model, project model, and affine model all exhibit equally strong results in terms of recall@10 and recall@20. However, the affine model outperforms the others, achieving the highest recall@1 and recall@5.

\begin{figure}[!t]
	\centering
	\includegraphics[width=3.5in]{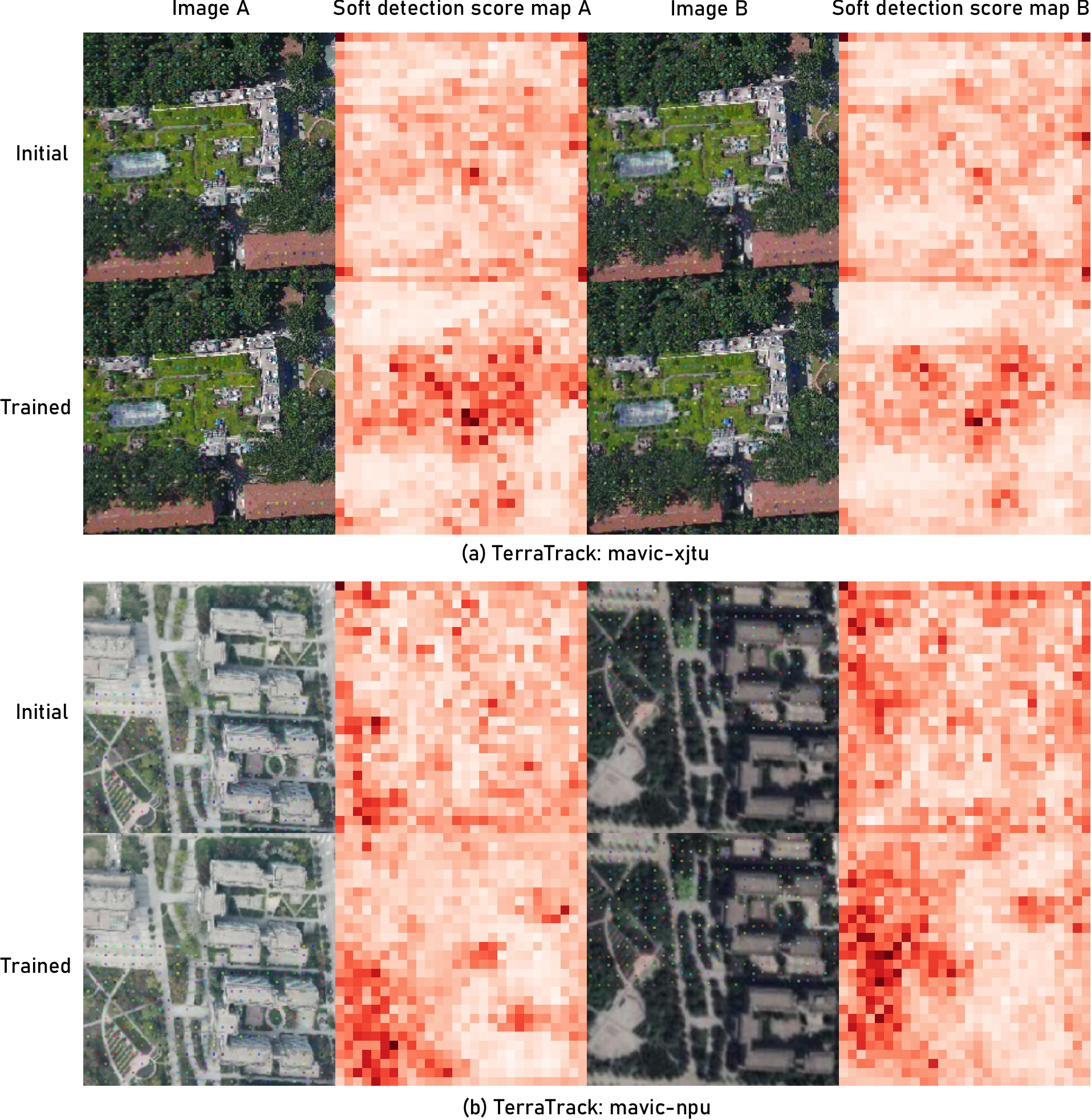}
	\caption{\textbf{Soft detection scores for different scenes before and after training.} White represents low soft-detection scores while red signifies higher ones. The contrast of the trained soft detection score map is significantly increased relative to its initial counterpart, which is more helpful in capturing the invariant information in the image.}
	\label{fig:softscores}
\end{figure}

Additionally, as shown in Fig. \ref{fig:softscores}, the contrast of the trained soft detection score map is significantly increased relative to its initial counterpart, which is more helpful in capturing the invariant information in the image.
\subsubsection{Pose Estimation and Localization}
The rerank top $5$ candidates and the corresponding matching keypoints pairs enable us to establish matching relationships between candidate images and query image. Solving this cross-domain stereo matching problem enable our CurriculumLoc to return a more accurate position closer to the query, which may not be present in the candidate database.

\begin{figure}[!t]
	\centering
	\includegraphics[width=3.5in]{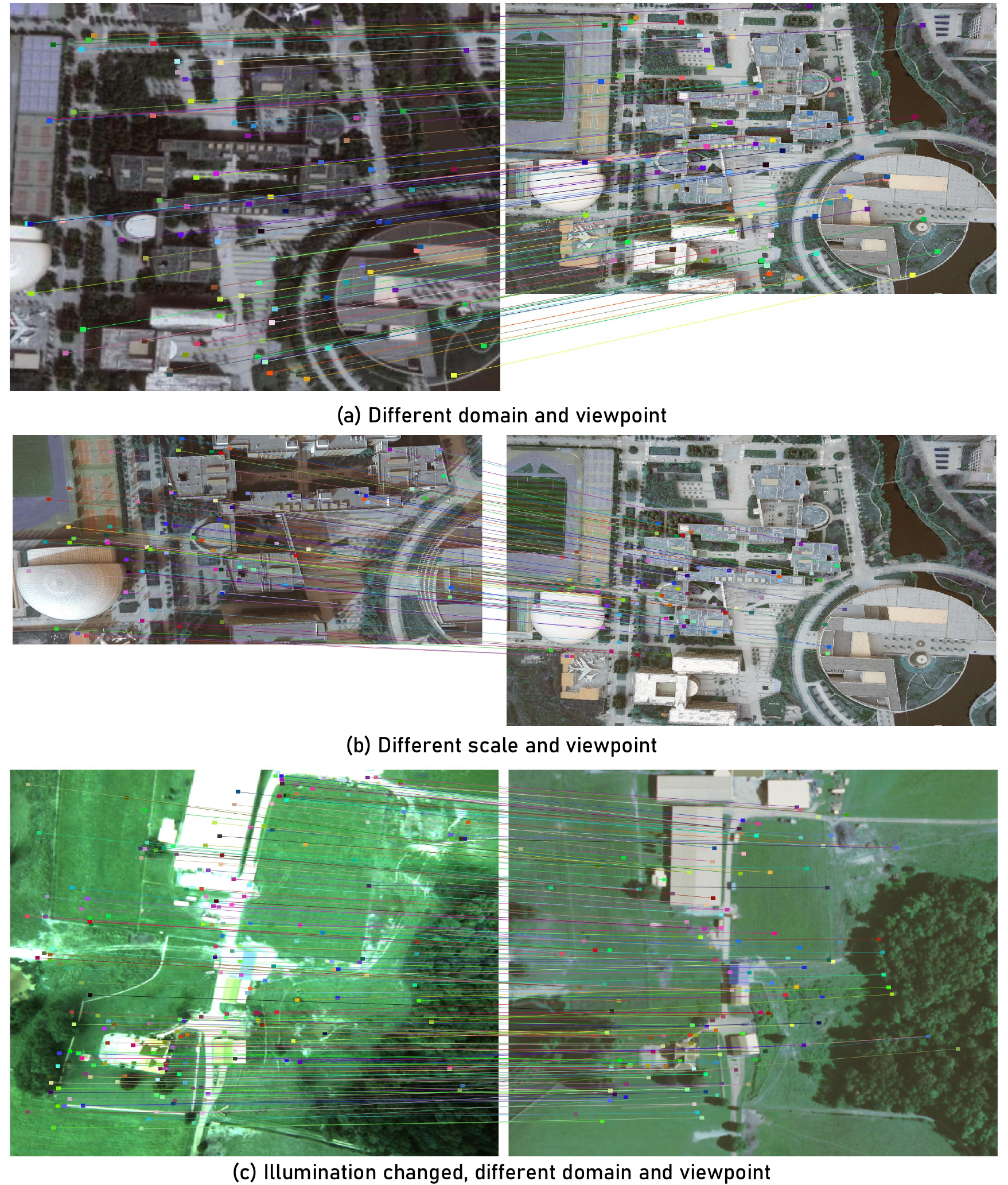}
	\caption{Final stereo matching results of matched image pairs from ALTO and TerraTrack datatsets in various challenge conditions.}
	\label{fig:match}
\end{figure}

\begin{figure}[!t]
	\centering
	\includegraphics[width=3.5in]{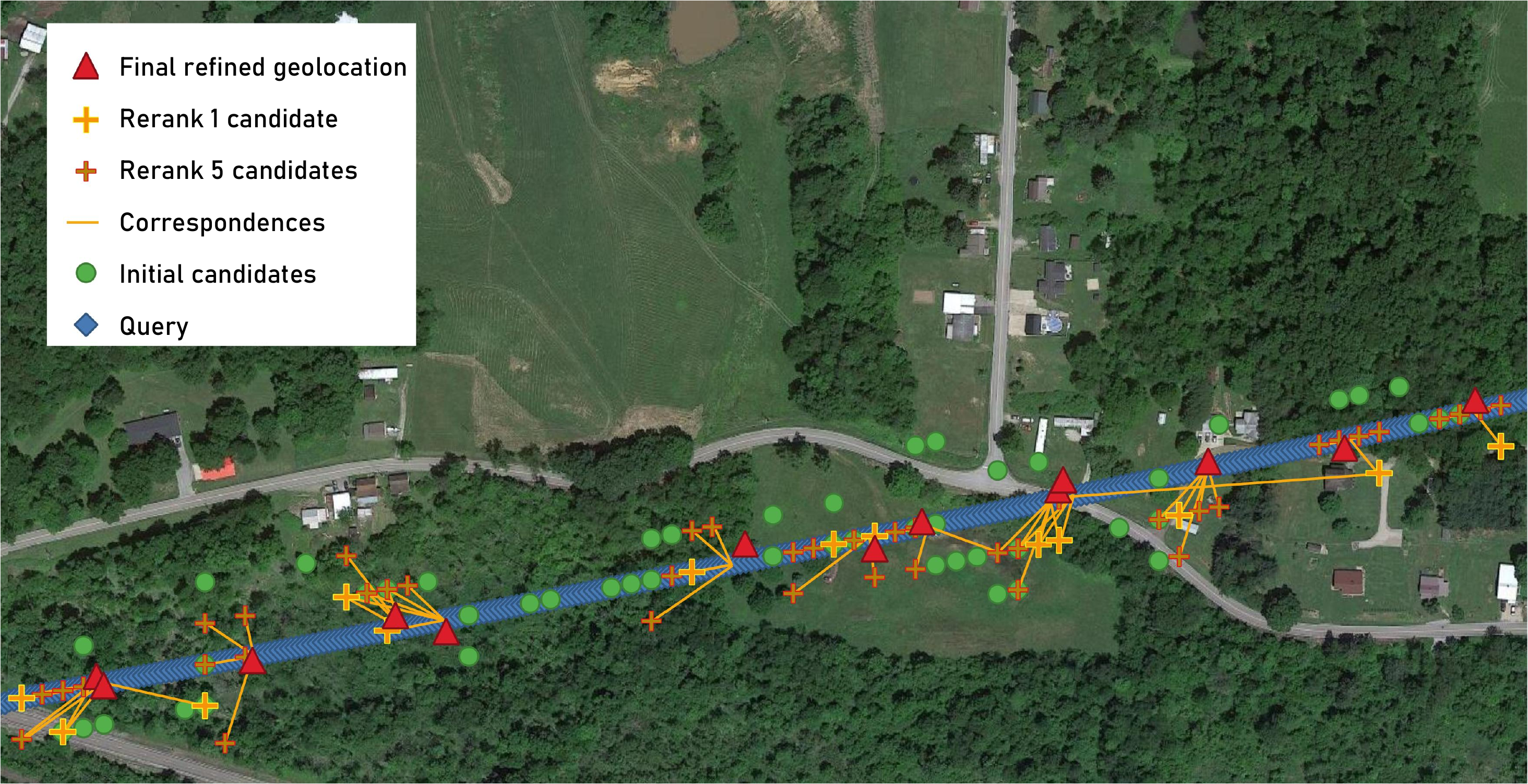}
	\caption{An overhead view of the trajectories generated by our CurriculumLoc is shown on the satellite map at ALTO. The green spot and orange cross are the location of the initial candidates (top-5) from global retrieval and refined candidates (top-5) by rerank. Red markers represent positions after PnP estimation. The orange line segment shows the correspondence between the rerank candidates and query. The blue trace represents the query position of the GPS.}
	\label{fig:localization-1}
\end{figure}

The visualization of the keypoint correspondences between images pairs from ALTO and TerraTrack dataset is illustrated in Fig. \ref{fig:match}.  Our \emph{Swin-Descriptors} detect and describe keypoints are robust to different domain and significant changes in scale, and in textureless regions, the form of our descriptor construction is able to identify correspondences between grass across different domain images.
Fig. \ref{fig:localization-1} illustrate the illustrates the improvement in localization accuracy from reranking to pose estimation and location refinement. It also shows the progress from the initial retrieval to the reranking stage.

\subsection{Localization under Challenging Conditions}
\begin{figure}[!t]
	\centering
	\includegraphics[width=3.5in]{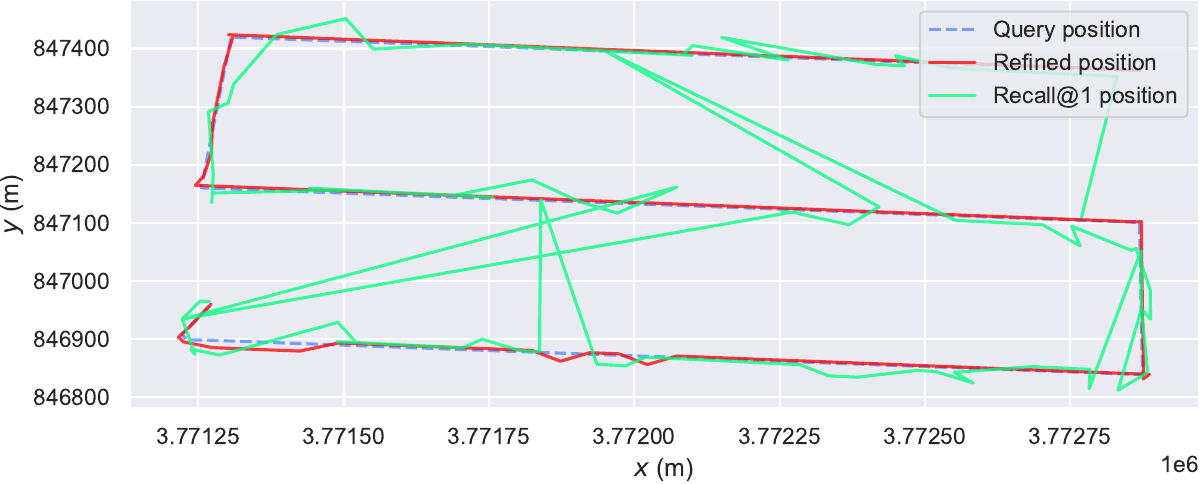}
	\caption{The trajectory of rerank recall@1 and final refined position.}
	\label{fig:traj}
\end{figure}

\begin{figure}[!t]
	\centering
	\includegraphics[width=3.5in]{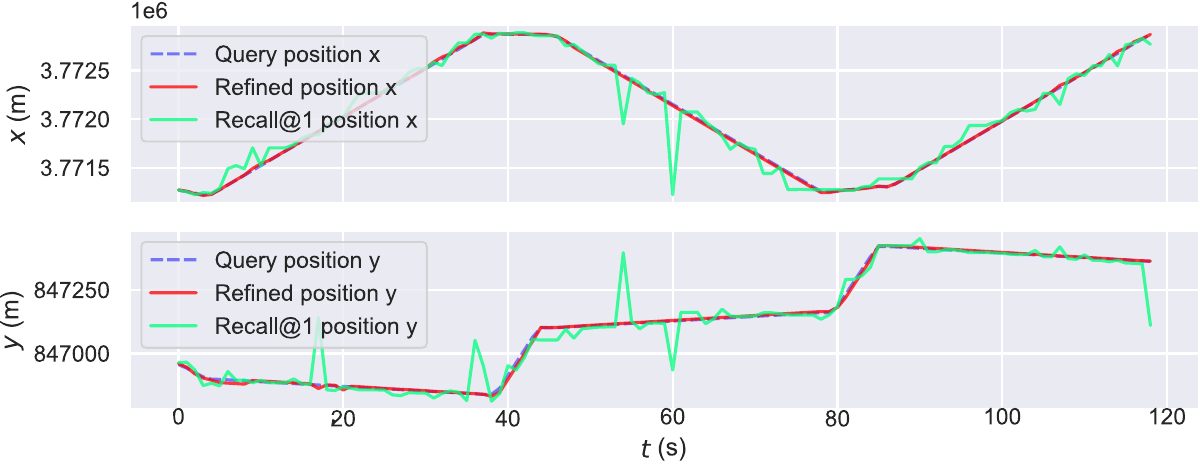}
	\caption{The trajectory of rerank recall@1 and final refined position in different axis.}
	\label{fig:xyz_traj}
\end{figure}

\begin{figure}[!t]
	\centering
	\includegraphics[width=3.5in]{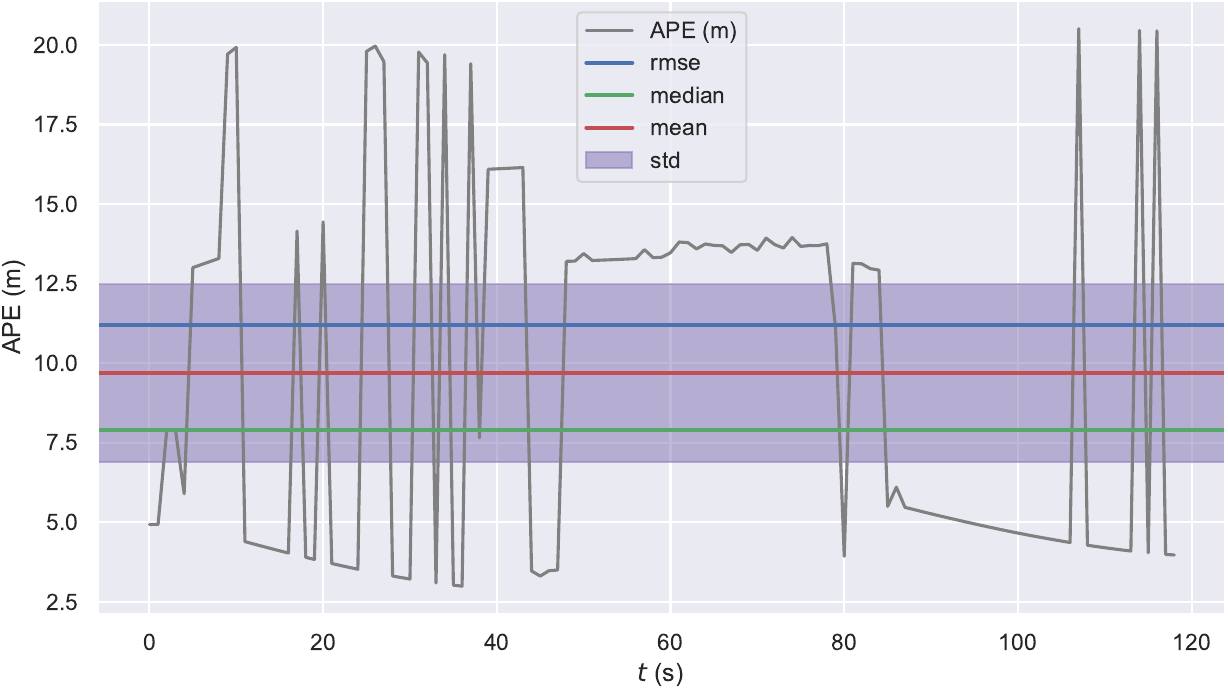}
	\caption{The absolute position error of final refined position.}
	\label{fig:ape_rmse_mean_std}
\end{figure}

The previous experiments showed that our approach through rerank outperforms comparable with the state-of-the-art retrieval methods and further improve the accuracy through pose estimation. In this experiment, we show that our approach is enough robust to be applied under a challenging condition: Localizing images under severe illumination changes and in different scale and domain.
%Fig. \ref{fig:npu} shows the overhead view of the final position trajectories generated by our CurriculumLoc is shown on the satellite map at TerraTrack (mavic-npu). 
Fig. \ref{fig:traj} and Fig. \ref{fig:xyz_traj} show the trajectories of the recall@1 position and the final refined position. In particular, Fig. \ref{fig:xyz_traj} displays the position of the axis of x, y, and the x-axis and y-axis represent east and north coordinates in UTM system, respectively. It can be seen that our method has well refined the recall@1 position results. 
The mean, std, median, rmse, and APE results of the final position by our \emph{CurriculumLoc} are shown in Fig. \ref{fig:ape_rmse_mean_std}. It can be seen that only using our visual geolocalization solution, the mean error of localization is less than 10$m$, while the relative flight height of this data is higher than 500 meters, and the std, median, rmse and APE results also demonstrate that our \emph{CurriculumLoc} is competent for geolocation of practical UAV applications. 
%
%\begin{figure}[!t]
%	\centering
%	\includegraphics[width=3.5in]{fig-final/npu_final_location_crop}
%	\caption{An overhead view of the trajectories generated by our CurriculumLoc is shown on the satellite map at TerraTrack (mavic-npu). The blue spot and red marker are the location of the query final refined position.}
%	\label{fig:npu}
%\end{figure}
%

\section{Conclusion}

Our designed \textit{Swin Descriptors} address the issue of context information loss caused by the limited receptive field of CNNs in dense feature extraction. It constructs an Encoder-Decoder transformer and enhances interaction between different windows through shift window attention in Swin Transformer. This enables the extraction of more robust and reliable invariant features under extreme weather conditions, viewpoint changes, and scale variations.
% which are crucial for accurate stereo matching.
%
Furthermore, our soft detection score map is detected from dense feature map and optimized by pixel geometric supervision, our keypoints is distinctive and corresponding descriptors enable fully enjoy global semantic awareness and local geometric verification.  We utilizing these ketpoints and descriptors to rerank and estimate a more precise position. Different from existing geo-localization algorithms that can only return the nearest neighboring geo-tag in the database, our algorithm provides an accurate global location.
Thorough experiments on retrieval-based localization, keypoint detection and matching, localization under challenging conditions, and ablation study validated that our \emph{Swin-Descriptors} can embrace both distinctive and invariance power and our\emph{CurriculumLoc} can outperforms existing cross-domian geolocalization methods. The future works include exploring more efficient backbone networks, as well as integrating our method into practical SLAM and other reconstruction applications.

\bibliography{citation}
\bibliographystyle{unsrt} 

\end{document}